%% file: main.tex
\documentclass[runningheads]{llncs}

\usepackage{eccv}

\usepackage{eccvabbrv}

\usepackage{graphicx}
\usepackage{booktabs}

\usepackage[accsupp]{axessibility}  % Improves PDF readability for those with disabilities.

\usepackage{hyperref}

\usepackage{orcidlink}

\usepackage{multirow}
\usepackage{multicol}
\usepackage[linesnumbered, boxed, ruled]{algorithm2e}
\usepackage{wrapfig}

\definecolor{lightred}{RGB}{251,49,153}
\begin{document}

% ---------------------------------------------------------------
% TODO REVIEW: Replace with your title
\title{\textsc{InstaStyle}: Inversion Noise of a Stylized Image is Secretly a Style Adviser} 

% TODO REVIEW: If the paper title is too long for the running head, you can set
% an abbreviated paper title here. If not, comment out.
\titlerunning{\textsc{InstaStyle}}

% TODO FINAL: Replace with your author list. 
% Include the authors' OCRID for the camera-ready version, if at all possible.
% \author{First Author\inst{1}\orcidlink{0000-1111-2222-3333} \and
% Second Author\inst{2,3}\orcidlink{1111-2222-3333-4444} \and
% Third Author\inst{3}\orcidlink{2222--3333-4444-5555}}

\author{
    Xing Cui$^1$ \quad 
    Zekun Li$^2$ \quad 
    Peipei Li$^1$\thanks{Corresponding author.} \quad 
    Huaibo Huang$^3$ \quad 
    Xuannan Liu$^1$ \quad 
    Zhaofeng He$^1$ \\
}
\institute{ Beijing University of Posts and Telecommunications \and
    University of California, Santa Barbara \and
    MAIS \& NLPR, Institute of Automation, Chinese Academy of Sciences \\
\email{ \{cuixing, lipeipei, liuxuannan, zhaofenghe\}@bupt.edu.cn}\\
\email{zekunli@cs.ucsb.edu} \quad
\email{ huaibo.huang@cripac.ia.ac.cn } \\
\textcolor{lightred}{\url{https://cuixing100876.github.io/instastyle.github.io/}}
}

% TODO FINAL: Replace with an abbreviated list of authors.
\authorrunning{X.~Cui et al.}
% First names are abbreviated in the running head.
% If there are more than two authors, 'et al.' is used.

% TODO FINAL: Replace with your institution list.
% \institute{Princeton University, Princeton NJ 08544, USA \and
% Springer Heidelberg, Tiergartenstr.~17, 69121 Heidelberg, Germany
% \email{lncs@springer.com}\\
% \url{http://www.springer.com/gp/computer-science/lncs} \and
% ABC Institute, Rupert-Karls-University Heidelberg, Heidelberg, Germany\\
% \email{\{abc,lncs\}@uni-heidelberg.de}}

\maketitle

\input{sec/0_abstract}    
\input{sec/1_intro}
\input{sec/2_related}
\input{sec/3_method}

\input{sec/4_experiment}

\input{sec/5_conclusion}

% \clearpage\mbox{}Page \thepage\ of the manuscript.
% \clearpage\mbox{}Page \thepage\ of the manuscript.
% \clearpage\mbox{}Page \thepage\ of the manuscript.
% \clearpage\mbox{}Page \thepage\ of the manuscript.
% \clearpage\mbox{}Page \thepage\ of the manuscript. This is the last page.
% \par\vfill\par
% Now we have reached the maximum length of an ECCV \ECCVyear{} submission (excluding references).
% References should start immediately after the main text, but can continue past p.\ 14 if needed.
% \clearpage  % TODO REVIEW/FINAL: This \clearpage needs to be removed from both review and camera-ready versions.

% ---- Bibliography ----
%
% BibTeX users should specify bibliography style 'splncs04'.
% References will then be sorted and formatted in the correct style.

\bibliographystyle{splncs04}
\bibliography{main}
% \printbibliography

\input{sec/X_suppl}

\end{document}

%% file: sec/0_abstract.tex
\begin{center}
    \includegraphics[width=0.9\linewidth]{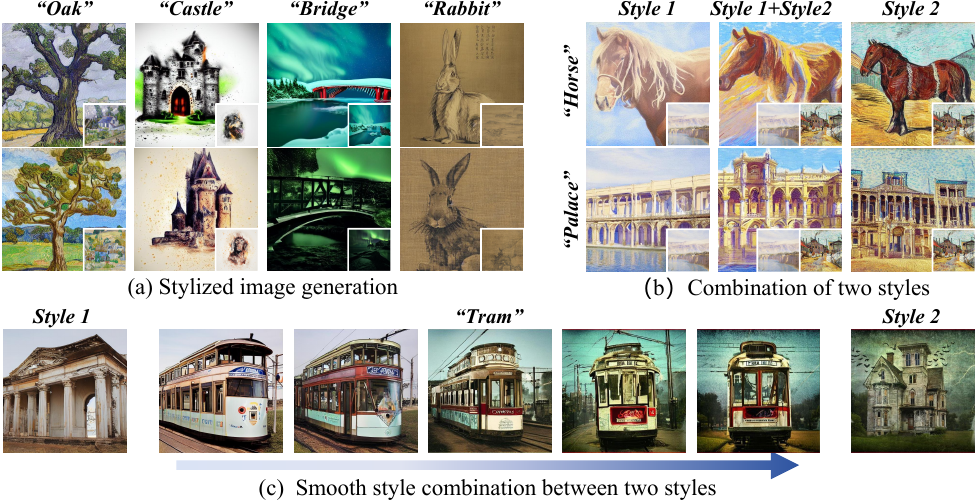}%
    \captionof{figure}{
    \textbf{Visualization of \textsc{InstaStyle}.} (a) Our method excels at capturing style details and distinguishing between similar styles. 
    (b) The first and third columns show images styled with reference to style 1 and style 2, respectively. The middle column shows images in a combined style.
    (c) Our method supports adjusting the degree of two styles during combination, dynamically ranging from one style to another.}
    \label{fig:figure1}
\end{center}%

\begin{abstract}
Stylized text-to-image generation focuses on creating images from textual descriptions while adhering to a style specified by reference images. However, subtle style variations within different reference images can hinder the model from accurately learning the target style. In this paper, we propose \textsc{InstaStyle}, a novel approach that excels in generating high-fidelity stylized images with only a single reference image. Our approach is based on the finding that the inversion noise from a stylized reference image inherently carries the style signal, as evidenced by their non-zero signal-to-noise ratio. We employ DDIM inversion to extract this noise from the reference image and leverage a diffusion model to generate new stylized images from the ``style'' noise.
Additionally, the inherent ambiguity and bias of textual prompts impede the precise conveying of style during image inversion. 
To address this, we devise prompt refinement, which learns a style token assisted by human feedback.
Qualitative and quantitative experimental results demonstrate that \textsc{InstaStyle} achieves superior performance compared to current benchmarks. Furthermore, our approach also showcases its capability in the creative task of style combination with mixed inversion noise. 
% Code is available at.
\keywords{Stylized image generation \and Inversion noise \and  Signal-to-noise ratio \and Prompt refinement}
\end{abstract}

%% file: sec/1_intro.tex
\section{Introduction}
\label{sec:intro}
\begin{wrapfigure}{r}{0.5\textwidth}
\vspace{-7mm}
    \centering
    \includegraphics[width=\linewidth]{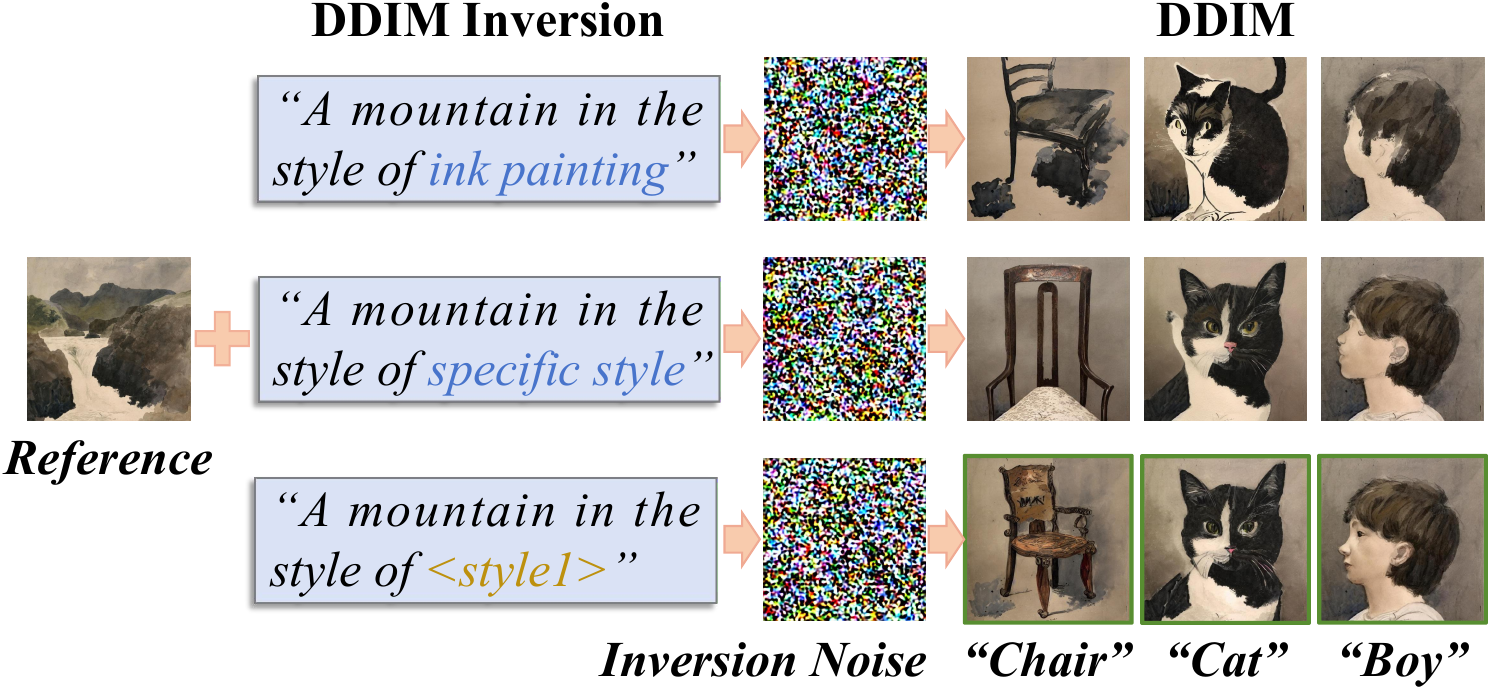}
    \caption{\textbf{Motivation.} Sampling from the inversion noise of a reference image can generate stylized images. However, the optimal style token varies for each case among \textcolor{blue}{human-written style tokens}. Our \textcolor{brown}{learnable style token}, \ie, ``\textless style1\textgreater'', shows greater universality across various scenarios.}
    \label{fig:figure-prompt}
\vspace{-4mm}
\end{wrapfigure}

Recently, the advent of DDPMs~\cite{ho2020denoising,sohl2015deep} has ushered in a new era of image generation.
Existing text-to-image generation methods~\cite{nichol2022glide,ramesh2021zero,rombach2022high,cui2023chatedit} can generate images in coarse-grained styles by incorporating style descriptions in the textual prompt.
However, the textual prompt is ambiguous which makes it hard to express style precisely~\cite{gal2022stylegan}.
Recently, personalized generation~\cite{gal2022image,wei2023elite,ruiz2023dreambooth} has been introduced to generate novel concepts, which can be utilized for stylized generation by viewing the reference style as a new concept. 
These approaches typically necessitate multiple images as references. However, the subtle style variations in these multiple images present a challenge for the model to accurately learn and replicate the intended style.
As shown in the first column of Fig.~\ref{fig:figure1} (a), although the reference image \textit{Houses at Auvers} (top) and \textit{Thatched Cottages by a Hill} (bottom) are all Van Gogh’s artworks, the differences in terms of color and texture~\cite{chen2021dualast} pose a significant challenge.
Some recent approaches~\cite{sohn2023learning,Cho_2023_ICCV,Xu_2023_ICCV} are designed to more accurately describe the target style during generation. However, they may ignore the fine-grained style information in the reference images.

In this paper, we propose \textbf{\textsc{InstaStyle}} (\textbf{I}nversion \textbf{N}oise of a \textbf{St}ylized Image is Secretly  \textbf{a} \textbf{Style} Adviser) based on the observation that the inversion noise from a stylized reference image inherently contains the style signal.
Specifically, our approach only requires a single reference image, which is transformed into a noise via DDIM inversion. The inversion noise, which preserves the style signal, is then used to generate stylized images by utilizing diffusion models.
To shed light on this phenomenon, we demonstrate that the inversion noise maintains a non-zero signal-to-noise ratio in Sec.~\ref{sec:non-zero}, indicating that it retains essential information (including the style signal) from the reference image.

Nevertheless, directly using human-written prompts to describe styles for inverting the reference image often encounters challenges due to the inherent ambiguity of natural language. 
For instance, specific style descriptors like ``ink painting'' don't always work effectively with various objects. As shown in Fig.~\ref{fig:figure-prompt}, the style ``ink painting'' might suit an object like a ``chair'' but not as well with a ``boy''. This inconsistency can arise because ``ink painting'' is typically associated with landscapes and might not yield optimal results for human subjects. Conversely, when using vague descriptors like ``specific style'', they may fail to provide enough information, leading to unpredictable generation quality (see the second row in Fig.~\ref{fig:figure-prompt}).
To address this, we propose \textit{Prompt Refinement} to learn a style token with the assistance of human feedback.
As illustrated in the last row of Fig.~\ref{fig:figure-prompt}, 
the style information is boosted by using the learned style token in the noising process to obtain a better ``style'' noise.
Therefore, our prompt refinement stands apart from previous methods~\cite{sohn2023learning,Cho_2023_ICCV,Xu_2023_ICCV}, where style tokens are utilized in the denoising process to generate target images.
During prompt refinement, we first collect images via human feedback. 
Then we optimize the embedding of the style token and the key and value projection matrices in the cross-attention layers of the diffusion model.

As shown in Fig.~\ref {fig:figure1} (a), our approach can effectively retain the fine-grained style in the reference image as well as generate new objects in high fidelity. Furthermore, our method supports the creative generation of style combination (Fig.~\ref {fig:figure1} (b) and (c)) and allows adjusting the degree of two styles dynamically (Fig.~\ref {fig:figure1} (c)).
To sum up, our contributions are as follows:
\begin{itemize}
    \item We find that the inversion noise of a reference image via DDIM inversion potentially retains style information and evidence this observation by the non-zero signal-to-noise ratio of the inversion noise. 
    \item We propose \textsc{InstaStyle} based on our observation, which leverages the style signal in inversion noise. Additionally, to better represent the style during inversion, a prompt refinement scheme is designed to learn the style token. Human feedback is also introduced to enhance model alignment with human preferences during prompt refinement. 
    \item  Qualitative and quantitative experimental results show the superiority of our \textsc{InstaStyle} in generating high-quality stylized images. Moreover, it exhibits promising potential in creative scenarios like style combinations.
\end{itemize}

%% file: sec/2_related.tex
\section{Related Work}
\label{sec:related}

\subsection{Text-to-image Synthesis}
Image synthesis is an essential subject in computer vision~\cite{dhariwal2021diffusion,meng2021sdedit,jia2022theme,tang2016tri}. Previous works are based on variational auto-encoder~\cite{kingma2013auto, li2023progressive} or generative adversarial networks~\cite{li2019global, cui2023chatedit,goodfellow2020generative, teng2023exploring, li2019m2fpa}.
With the advance of pretrained vison-language models~\cite{radford2021learning,wang2023learning} and diffusion models~\cite{song2020denoising,zhang2023towards,li2023pluralistic}, text-to-image generation~\cite{esser2021taming,nichol2022glide,ramesh2021zero,cui2024localize,wang2024stablegarment} has been widely studied and shown remarkable generalization ability. 
Recently, Lin \etal~\cite{lin2023common} have pointed out that the noise of the last step during noising in the training process has a non-zero signal-to-noise ratio (SNR), \ie, there exists a signal leak. This results in a misalignment between the training and inference process. Therefore, some works~\cite{lin2023common, salimans2021progressive} propose to train diffusion models by enforcing the SNR to zero to avoid the signal leak. On the contrary, our approach leverages the leaked signal, which potentially includes the style details, from the reference image for the stylized generation.

As these methods ignore the concepts that do not appear in the training set~\cite{gal2022image}, some works~\cite{gal2022image,ruiz2023dreambooth,ruiz2023hyperdreambooth} study personalized text-to-image generation which aims to adapt text-to-image models to new concepts given several reference images.
For example, Textual Inversion~\cite{gal2022image} introduces and optimizes a word vector for each new concept.
Subsequent works~\cite{wei2023elite,zhou2023enhancing,alaluf2023neural} further enhance the flexibility and adaptiveness of the learning strategy.
Some methods~\cite{hu2021lora,ruiz2023dreambooth,ruiz2023hyperdreambooth} propose to finetune the original diffusion model, showing more satisfactory results. 
% For example, DreamBooth~\cite{ruiz2023dreambooth} finetunes all the parameters in the pretrained model. 
However, maintaining fine-grained style while simultaneously generating new objects remains a challenge for current personalized generation methods~\cite{sohn2023styledrop}.

\subsection{Stylized Image Generation}
Stylized image generation is a new paradigm for image generation which aims to generate content in a specific style given a few reference images. Although it is similar to the neural style transfer task~\cite{gatys2016image,gatys2017controlling,huang2017arbitrary,lin2021drafting,wang2020collaborative,karras2019style} which generates stylized images as well, they are fundamentally different. Style transfer solves an image translation task, focusing on translating a content image to target style~\cite{jing2020dynamic,Yang_2023_ICCV,Wang_2023_ICCV,Gu_2023_ICCV,Ke_2023_CVPR,Xu_2023_CVPR,Huang_2023_CVPR,Wen_2023_CVPR,Xie_2022_CVPR}. For example, some works explore the global and local information to preserve the content of the source image~\cite{Hong_2023_ICCV,Zhu_2023_ICCV,deng2022stytr2,wu2021styleformer,Zhang_2023_CVPR,Tang_2023_CVPR,jing2022learning,jing2018stroke}.
On the contrary, stylized image generation is geared towards generating a new image in specific style conditioned on a text.
For example, ZDAIS~\cite{sohn2023learning} views stylized image generation as a domain adaptive task, where each style belongs to a domain. It learns disentangled prompts to adapt the model to new domains. 
Some approaches~\cite{sohn2023styledrop,Everaert_2023_ICCV} fine-tune the model to capture style properties. 
Our work differs from theirs by proposing a framework based on our observation that inversion noise contains style information.
% Our work differs from these previous works. Firstly, we reveal that the inversion noise of the reference image can act as a style adviser to provide fine-grained style information. Secondly, we avoid the ambiguity caused by the natural language prompt by learning a style token for better image inversion and generation. 
% Moreover, benefiting from the inversion ``style'' noise and the style token, our method supports the dynamic combination of multiple styles. 

%% file: sec/3_method.tex
\section{Method}
\label{sec:method}
\begin{figure*}[t]
\begin{center}
\includegraphics[width=\linewidth]{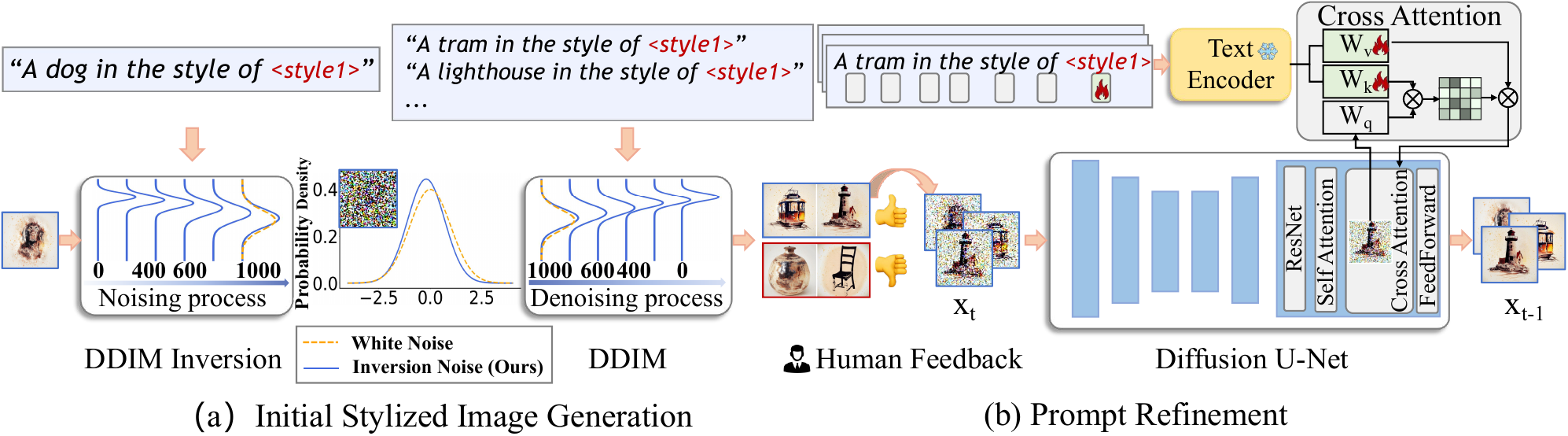}
\end{center}
\caption{\textbf{The training process of \textsc{InstaStyle}.} (a) The first stage is initial stylized image generation. The reference image is inverted to noise conditioned on a prompt via DDIM Inversion. Then the inversion noise is utilized to generate initial stylized images. (b) The second stage is prompt refinement which leverages the selected high-quality initial stylized images to learn a style token. }\label{fig:framework}
\end{figure*}

As shown in Fig.~\ref{fig:framework}, our proposed method involves two main stages. In the first stage, we employ DDIM inversion to transform the reference image into noise. Notably, the inversion noise exhibits a non-zero signal-to-noise ratio, suggesting the presence of style signals from the reference image.
Subsequently, we generate $M$ stylized images from the inversion noise conditioned on the given textual prompts.
Due to the inherent ambiguity of the textual prompts that describe the style, it is challenging to precisely convey the desired style. Addressing this, the second stage involves the incorporation of human feedback to select $N$ high-quality generated images from the first stage. The selected images are then used to learn a style token via prompt refinement.

\subsection{Preliminaries}\label{sec:preliminaries}
We begin by reviewing the fundamental diffusion model. Subsequently, we introduce Stable Diffusion~\cite{rombach2022high}, a key component of our framework. Finally, we discuss both DDIM and DDIM inversion. In our method, DDIM serves the purpose of denoising a noise to generate new images. Concurrently, DDIM inversion is utilized to transform the reference image into a corresponding ``style'' noise.

\noindent\textbf{Diffusion models.}
Diffusion models~\cite{ho2020denoising,sohl2015deep} contain a forward process and a backward process. The forward process adds noise to the data according to a predetermined, non-learned variance schedule $\beta _{1},...,\beta _{T}$:
\begin{equation}
    q(z_t|z_{t-1}):=\mathcal{N}(z_t;\sqrt{1-\beta_t}z_{t-1},\beta_t\mathbf{I}).
\end{equation}
An important property of the forward process is that we can obtain the latent variable $z_t$ directly based on $z_0$:
\begin{equation}
    \label{eq:qt_q0}
    z_t=\sqrt{\alpha_t}z_0+\sqrt{1-\alpha_t}\varepsilon,
\end{equation}
where $\varepsilon\sim\mathcal{N}(\mathbf{0},\mathbf{I})$, $\alpha_t:=\prod_{i=1}^t(1-\beta_i)$.
Diffusion models restore the information by learning the backward process:
\begin{equation}
    p_\theta(z_{t-1}|z_t):=\mathcal{N}(z_{t-1};\mu_{\theta}(z_{t},t),\Sigma_{\theta}(z_{t},t)).
\end{equation}

\noindent\textbf{Stable Diffusion.}
In Stable Diffusion, the diffusion process happens in latent space~\cite{rombach2022high}. 
It utilizes a pretrained autoencoder which consists of an encoder and a decoder.
The encoder $\mathcal{E}(\cdot)$ maps an image $x$ into a latent code $z$ and the decoder $\mathcal{D}(\cdot)$ transforms the latent code back into an image. The denoising model $\varepsilon_\theta(\cdot)$ is an Unet~\cite{ronneberger2015u}.
During training, artificial noise is added to the sampled data $z_0$ according to timestamp $t$ based on~\cref{eq:qt_q0}, resulting in a noised sample $z_t$. Denoising model starts with $z_t$ and is trained to predict the artificial noise:
\begin{equation}
    \min_{\theta}E_{z_0,\varepsilon\sim N(0,I),t\sim\mathrm{Uniform}(1,T)}\left\|\varepsilon-\varepsilon_\theta(z_t,t,\mathcal{C})\right\|_2^2,
    \label{eq:sd_loss}
\end{equation}
where $\mathcal{C}=\psi(\mathcal{P})$ is the embedding of text condition $\mathcal{P}$.

\noindent\textbf{DDIM.}
In inference time, given a noise vector $z_T$, the noise is gradually removed by sequentially predicting the added noise for $T$ steps. DDIM~\cite{song2020denoising} is one of the denoising approaches with a deterministic process:
\begin{equation}
\label{eq:ddim}
\resizebox{0.7\textwidth}{!}{
    $z_{t-1}=\sqrt{\frac{\alpha_{t-1}}{\alpha_{t}}}z_{t}+\left(\sqrt{\frac{1}{\alpha_{t-1}}-1}-\sqrt{\frac{1}{\alpha_{t}}-1}\right)\cdot\tilde{\varepsilon}_\theta$,
    }
\end{equation}
where $\tilde{\varepsilon}_\theta$ is the estimated noise.

DDIM inversion~\cite{song2020denoising,mokady2023null} transforms an image to a noise conditioned on a prompt. The diffusion process is performed in the reverse direction, \ie, $z_0\to z_T$:
\begin{equation}
\label{eq:ddim_inverion}
\resizebox{0.7\textwidth}{!}{
    $z_{t+1}=\sqrt{\frac{\alpha_{t+1}}{\alpha_{t}}}z_{t}+\left(\sqrt{\frac{1}{\alpha_{t+1}}-1}-\sqrt{\frac{1}{\alpha_{t}}-1}\right)\cdot\tilde{\varepsilon}_\theta$.
    }
\end{equation}

\subsection{Initial Stylized Image Generation}\label{sec:non-zero}
As shown in Fig.~\ref{fig:framework} (a), the first stage of our \textsc{InstaStyle} involves obtaining the inversion noise via DDIM inversion and sampling images via DDIM.

In DDIM inversion, the added noise for each step is calculated conditioned on a prompt and gradually incorporated into the reference image following~\cref{eq:ddim_inverion}. 
Specifically, we set the learnable style token as a human-written style description in this stage.
We demonstrate that the inversion noise from a stylized reference image inherently carries the style signal from the perspective of the signal-to-noise ratio. As the estimated noise $\varepsilon_{\theta}(z_{t},t,\mathcal{C})$ is trained to approximate the artificial noise $\varepsilon\sim\mathcal{N}(\mathbf{0},\mathbf{I})$, we assume that $\varepsilon_{\theta}(z_{t},t,\mathcal{C})\sim\mathcal{N}(\mathbf{0},\mathbf{I})$.
According to~\cref{eq:ddim_inverion}, $z_{t+1}$ can be approximated in a closed form:
\begin{equation}
\resizebox{0.8\textwidth}{!}{
    $z_{t+1}=\sqrt{\frac{\alpha _{t+1}}{\alpha _0}}z_0+\sqrt{\sum_{i=0}^t{\frac{\alpha _{t+1}}{\alpha _{i+1}}\left( \sqrt{\frac{1}{\alpha _{i+1}}-1}-\sqrt{\frac{1}{\alpha _i}-1} \right) ^2}} \cdot \bar{\varepsilon}_{0},$
    }
\end{equation}
where $\bar{\varepsilon}_{0}\sim\mathcal{N}(\mathbf{0},\mathbf{I})$. 

\noindent\textbf{Signal-to-Noise ratio.}
Signal-to-noise ratio (SNR) is introduced to measure the ratio of signals from the original image preserved in the noise~\cite{rombach2022high,lin2023common}. The SNR of the inversion noise can be calculated as:
\begin{equation}
\resizebox{0.6\textwidth}{!}{
    $\mathrm{SNR}(t):=\frac{1}{\sum_{i=0}^t{\frac{\alpha _{0}}{\alpha _{i+1}}\left( \sqrt{\frac{1}{\alpha _{i+1}}-1}-\sqrt{\frac{1}{\alpha _i}-1} \right) ^2}}.$
    }
\end{equation}
We present detailed derivations in Sec.~\ref{sup:snr} in Supplementary Material. 
In Stable Diffusion~\cite{rombach2022high}, the predetermined variance schedule $\beta_{t}=(\sqrt{0.00085}\cdot(1-j)+\sqrt{0.012}\cdot j)^{2}$, where $j=\frac{t-1}{T-1}$. At the terminal timestep $T=1000$, the $SNR(T)=0.015144$, \ie, $z_T$ has a none-zero signal-to-noise ratio.
This non-zero signal-to-noise ratio suggests that the noise obtained via DDIM inversion still retains information from the reference image and deviates from white noise. 
Besides, we conduct qualitative experiments and observe that style information can consistently be preserved at the terminal timestep (refer to Sec.~\ref{sup:analysis_DDIM} in Supplementary Material).
Therefore, we can generate target stylized images by sampling from the inversion noise based on~\cref{eq:ddim}.

\subsection{Prompt Refinement}\label{sec:prompt_refinement}
As the natural language prompts may not precisely convey the style (Fig.~\ref{fig:figure-prompt}), we propose Prompt Refinement to learn the style tokens as shown in Fig.~\ref{fig:framework} (b). 
We utilize the generated stylized images in the first stage to constitute the training data for prompt refinement learning. While these generated images may not always be precise, we observe that most of them successfully retain the style information alongside appropriate content. We manually select images that distinctly embody the reference style and the target object to build the dataset.

Specifically, we introduce a new token, \ie, ``\textless style1\textgreater'' to represent the style descriptor and learn its embedding. In practice, we initialize the new token with the embeddings of the textual style descriptor of the reference image.
The text condition is input to the model via two projection matrices in the cross-attention block of diffusion model  $\varepsilon_\theta$, \ie, $W^{k}\in\mathbb{R}^{d\times d'}$ and $W^{v}\in\mathbb{R}^{d\times d'}$.
Therefore, we also fine-tune these two projection matrices. 
The text feature $\mathbf{c}\in\mathbb{R}^{s\times d}$ is projected to key $K=\mathbf{c}W^{k}$ and value $V=\mathbf{c}W^{v}$. The query matrix $W^{q}\in\mathbb{R}^{l\times d'}$ projects the latent image feature $\mathbf{f}\in\mathbb{R}^{(h\times w)\times l}$ into query feature $Q=\mathbf{f}W^{q}$. Then the cross-attention~\cite{vaswani2017attention} is calculated as:
\begin{equation}
    \text{Attention}(Q,K,V)=\text{Softmax}\Big(\frac{QK^T}{\sqrt{d'}}\Big)V.
\end{equation}

We utilize the LoRA~\cite{hu2021lora} for model fine-tuning. Specifically, for a projection matrix $W\in\mathbb{R}^{d\times d'}$ (\ie, $W=W^k$ or $W=W^v$) to be fine-tuned, we update a low-rank residual rather than directly fine-tuning $W$. Formally, we denote the fine-tuned projection matrix as $W'=W+B A$, where $B \in \mathbb{R}^{d \times r}$, $A \in \mathbb{R}^{r \times d'} $, and the rank $ r \ll \min (d, d')$. During training, only $A$ and $B$ are learnable. For $y=Wx$, the modified forward pass is $y=Wx+BAx$. 
% Finally, our training objective is the same as that in Stable Diffusion~\cite{rombach2022high}, which is shown in~\cref{eq:sd_loss}. 
Putting the two stages together, our full algorithm is shown in~\cref{code}.

\begin{algorithm}[t]
\SetAlgoLined
\textbf{Input:}A source prompt embedding $\mathcal{C}=\psi(\mathcal{P})$ and a reference image $\mathcal{I}$. Initial style token embedding $v$ is included in the prompt embedding $\mathcal{C}$.\\ 
\textbf{Output:}  Optimized style token embedding $v^*$ and diffusion model $\varepsilon^*_\theta$.\\
\vspace{1mm} \hrule \vspace{1mm}
\tcp{\small\hspace{-2mm} Initial Stylized Image Generation}
Compute the inversion noise  $z_T$ using DDIM inversion over $\mathcal{I}$ conditioned on $\mathcal{C}$ based on~\cref{eq:ddim_inverion}; \\
Generate $M$ images conditioned on a predefined prompt set based on~\cref{eq:ddim};\\
\tcp{\small\hspace{-2mm} Prompt Refinement}  
Select $N$ images from $M$ images with human feedback as the training dataset $q(\mathbf{x}_{0},c)$;\\
\Repeat{converged}{
    $\mathbf{x}_{0},c\sim q(\mathbf{x}_{0},c)$;\\
    $z_0=\mathcal{E}(\mathbf{x}_{0})$;\\
    $t\sim $Uniform$( \{ 1, \ldots , T\} ) $;\\
    $\varepsilon\sim\mathcal{N}(\mathbf{0},\mathbf{I})$;\\
    $z_t=\sqrt{\alpha_t}z_0+\sqrt{1-\alpha_t}\varepsilon$;\\
    Take gradient step on $\nabla_{\theta,v} \left\|\varepsilon-\varepsilon_\theta(z_t,t,c)\right\|_2^2$;
}
\textbf{Return} $v^*$, $\varepsilon^*_\theta$

\caption{\textsc{InstaStyle}}
\label{code}
\end{algorithm}

\subsection{Inference}\label{sec:inference}
\noindent\textbf{Stylized image generation.}
For stylized image generation, we first learn the style token ``\textless style1\textgreater'' to describe the style in the reference image. During inference, the learned style token is integrated into the prompt that inverts image and the prompt that generates the target image. Besides, the diffusion model $\varepsilon_\theta$ is utilized as the backbone of DDIM inversion and DDIM. Specifically, the reference image is first inverted to a noise $z_T$ via DDIM inversion conditioned on a prompt containing the learned style token. Then we sample from the inversion noise conditioned on a target prompt which contains the target content and the learned style token via DDIM. 
Specifically, the estimated noise is calculated based on the classifier-free guidance~\cite{ho2021classifier}:
\begin{equation}
 \resizebox{0.8\textwidth}{!}{
$\tilde{\varepsilon}_\theta(z_t,t,\mathcal{C},\varnothing) =\varepsilon_\theta(z_t,t,\varnothing) 
 +w\cdot\bigl (\varepsilon_\theta(z_t,t,\mathcal{C})-\varepsilon_\theta(z_t,t,\varnothing) \bigr ),$
 }
\end{equation}
where $\varnothing=\psi($``''$)$ is the embedding of a null text.
$\varepsilon_\theta(z_t,t,\mathcal{C})$ represents the conditional predictions.
$w$ is the guidance scale parameter.

\noindent\textbf{Combination of two styles.} 
To combine two styles, we learn style tokens for each style, \ie, ``\textless style1\textgreater'' and ``\textless style2\textgreater''.
Specifically, we use the selected images of both styles to constitute the training set. Then we jointly optimize the embeddings of the style tokens and fine-tune the key projection and value projection in the cross-attention block.
During inference, we individually transform two reference images into their corresponding inversion noise $z_{t1}\in\mathbb{R}^{H\times W\times C}$ and $z_{t2}\in\mathbb{R}^{H\times W\times C}$ via DDIM inversion conditioned on a prompt containing their learned style token. 
As the style can be described by both the ``style'' noise and style token, we combine the styles from these two perspectives.
For the ``style'' noise, the combined inversion noise $z_t$ is obtained based on a masking strategy:
\begin{equation}
    z_t=(\mathbf{1}-\mathbf{M})\odot z_{t1}+\mathbf{M}\odot z_{t2},
\end{equation}
where $\odot$ is element-wise multiplication and $\mathbf{1}$ is a binary mask filled with ones.
$\mathbf{M}\in\{0,1\}^{H\times W}$ denotes a random binary mask indicating where to drop out and fill in from two inversion noises. The noise mix ratio between two inversion noises is $\alpha$, representing the percent of $\mathbf{M}$ that is set to 1 (the rest is set to 0). 
Finally, we utilize a composed guidance mechanism~\cite{liu2022compositional} to estimate the noise:
\begin{equation}
\resizebox{0.7\textwidth}{!}{
$\begin{aligned}
\tilde{\varepsilon}_\theta(z_t,t,\mathcal{C},\varnothing) & =\varepsilon_\theta(z_t,t,\varnothing)\\
& +w\cdot \bigl ( 1-\beta \bigr )\cdot \bigl (\varepsilon_\theta(z_t,t,\mathcal{C}_1)-\varepsilon_\theta(z_t,t,\varnothing)  \bigr )\\
&+w\cdot  \beta \cdot \bigl (\varepsilon_\theta(z_t,t,\mathcal{C}_2)-\varepsilon_\theta(z_t,t,\varnothing)  \bigr ) ,
\end{aligned}$}
\end{equation}
where $\beta$ is a prompt mix ratio. $\mathcal{C}_1$ and $\mathcal{C}_2$ are the embeddings of the target prompts for the two styles, respectively.
Denote the target object as \textless  obj\textgreater, they can be formulated as
$\mathcal{C}_1=\psi($``A \textless obj\textgreater \ in the style of \textless style1\textgreater''$)$ and
$\mathcal{C}_2=\psi($``A \textless obj\textgreater \ in the style of \textless style2\textgreater''$)$.

%% file: sec/4_experiment.tex
\section{Experiment}
\label{sec:experiment}
\subsection{Experimental Setting}

We collect 60 images as the reference style dataset. The complete information on these images is listed in~\cref{tab:supp_image_source} in the Supplementary Material.
Our method is implemented using PyTorch and executed on a single NVIDIA GeForce RTX 3090 GPU. The training procedure consists of 500 iterations, employing the Adam optimizer with a learning rate of $1 \times 10^{-5}$. During sampling, we set the guidance scale $w=2.5$. 
The number of generated images $M$ in the first stage is 15 whose details are shown in Sec.~\ref{sup: implementation_detail} in the Supplementary Material. The number of selected images $N$ in the second stage is set to $5$.

\begin{figure*}
\begin{center}
\includegraphics[width=\linewidth]{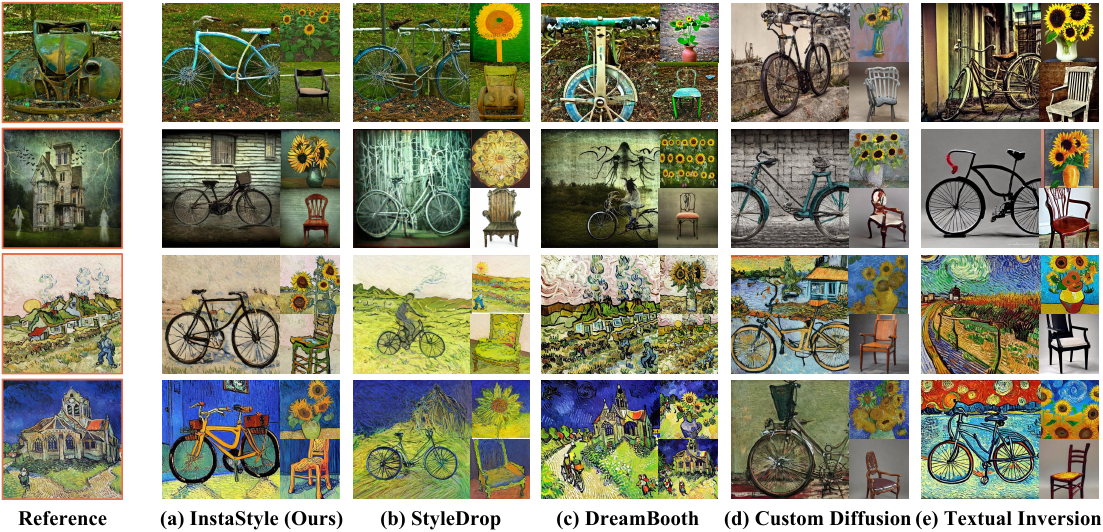}
\end{center}
\caption{\textbf{Qualitative comparison of stylized image generation on various styles.}  Objects for synthesis
are \textit{Bicycle}, \textit{Sunflowers} and \textit{Chair}. Our method excels at capturing fine-grained style information, such as color, textures, and brushstrokes.}\label{fig:figure_compare}
\end{figure*}

\subsection{Stylized Image Synthesis}
We conduct comparisons of \textsc{InstaStyle} against four recent methods following the open-sourced code$\footnote{\url{ https://github.com/aim-uofa/StyleDrop-PyTorch}}$$^{,}$$\footnote{\url{ https://github.com/huggingface/diffusers}}$, \ie, StyleDrop~\cite{sohn2023styledrop}, DreamBooth~\cite{ruiz2023dreambooth}, Custom Diffusion~\cite{kumari2023multi}, and Textual Inversion~\cite{gal2022image}.
Besides, following \cite{sohn2023learning}, we also make a comparison with style transfer methods~\cite{liu2021adaattn,zhang2022domain,wu2022ccpl,deng2022stytr2,Huang_2023_CVPR,hong2023aespa} to further illustrate the superiority of our approach.
Notably, the style transfer task takes images as content input,  in contrast to our task which only utilizes text prompts for content. To make a comparison between the two tasks, we additionally generate content images based on text prompts for style transfer methods.
Although we can compare the two tasks under this setting, the comparison is unfair to our approach because the content images utilized by style transfer methods may provide additional content information compared to our text prompts.

\begin{figure}[t]
\begin{center}
\includegraphics[width=\linewidth]{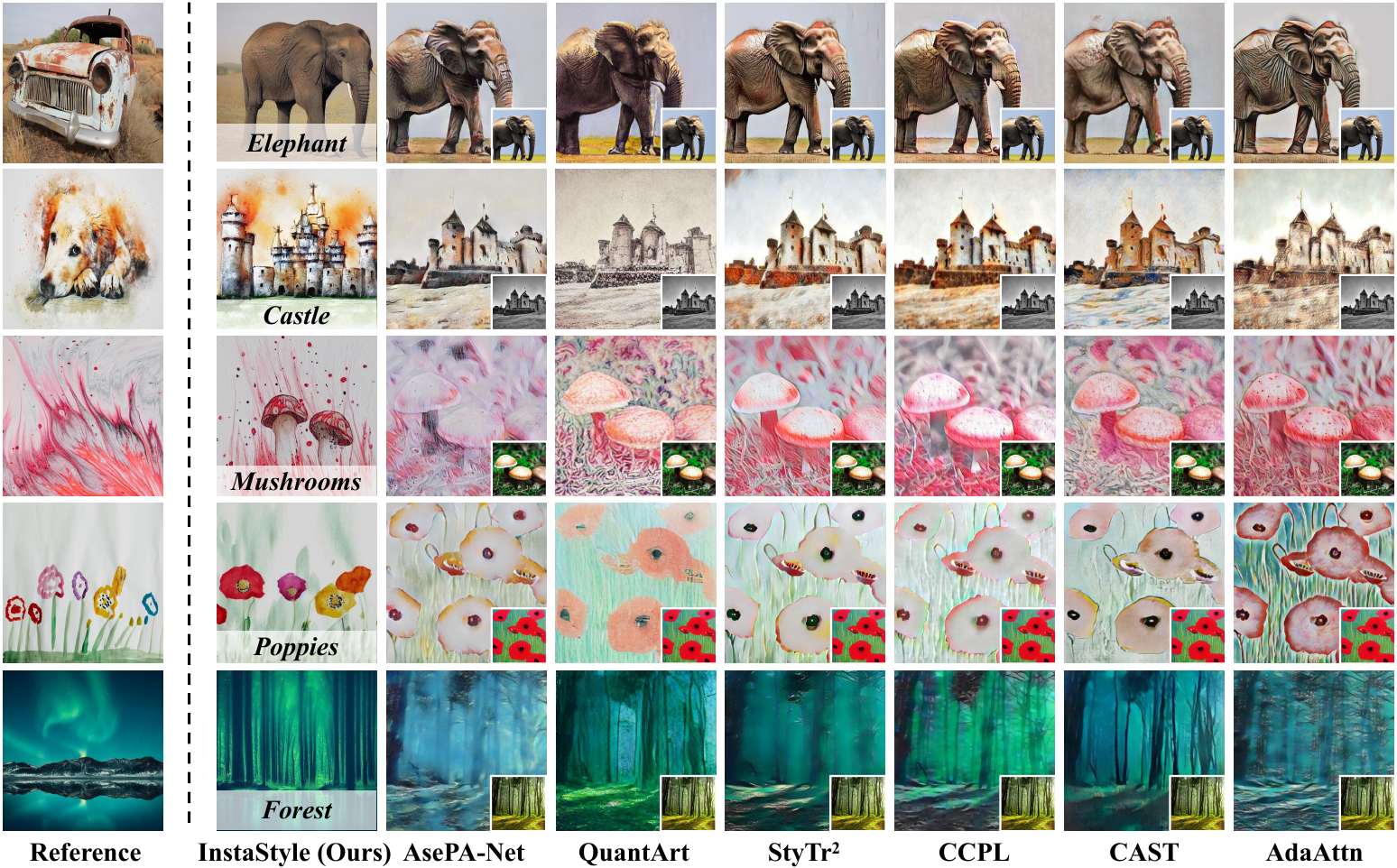}
\end{center}
\caption{\textbf{Qualitative comparison with style transfer methods.} For style transfer methods, content images are employed (shown in the bottom right). Despite our method relying solely on text for content, we achieve comparable performance in the fidelity of the content. Furthermore, we excel in preserving style details.} \label{fig:figure_styletransfer_compare}
\end{figure}

\noindent\textbf{Qualitative results.}
As shown in Fig.~\ref{fig:figure_compare}, textual Inversion and Custom Diffusion often yield an unsatisfactory style. 
Taking the well-known Van Gogh painting style (\eg, row 3 and row 4) as an example, they fail to capture fine-grained style information in the reference image.
The fine-tuning-based methods, \ie, DreamBooth and StyleDrop, result in distorted content. For example, they generate bicycles with leaked content from the reference image in the last row. 
In contrast, benefiting from the style signal in the inversion noise, our \textsc{InstaStyle} can generate stylized images with fine-grained style details and higher fidelity. Take the bicycle in~\cref{fig:figure_compare} as an example, our method can better preserve the color, textures, and brushstroke. Besides, the generated bicycle is more accurate.
\cref{fig:figure_styletransfer_compare} presents the qualitative comparisons with style transfer methods. 
Despite the inherent challenge of utilizing text as content input in contrast to style transfer methods that use the content image as input, our \textsc{InstaStyle} demonstrates comparable performance in content fidelity. Additionally, our \textsc{InstaStyle} exhibits superior performance in preserving the style details of the reference image, notably in terms of color and brushstroke characteristics.
More qualitative results are shown in~\cref{fig:sup_figure_elephant_tank,fig:sup_figure_vangogh_and_water,fig:sup_figure_compare_a_b,fig:sup_figure_compare_c_d,fig:sup_figure_stytransfer} in the Supplementary Material.

\begin{table}[t]
\caption{(a) Quantitative comparison regarding the style consistency score and content fidelity score. (b) User study. The results show the percentage of votes where the comparison method is preferred over ours, ties with ours and is inferior to ours. We use the abbreviation (ST) to represent approaches for style transfer tasks.}
\label{tab:comparison_user}
\vspace{-4mm}
    \begin{minipage}[t]{.5\linewidth}
      \centering
        \subcaption{}
        \resizebox{0.9\linewidth}{!}{%
        \setlength{\tabcolsep}{2mm}{
        \begin{tabular}{lcc}
        \toprule
        \textbf{Method} &Style ($\uparrow$) &Content ($\uparrow$)\\
        
        \midrule
        
        (ST) AdaAttN~\cite{liu2021adaattn} &0.587 &0.287\\
        (ST) CCPL~\cite{wu2022ccpl} &0.542 &0.286\\
        
        (ST) StyTr$^{2}$~\cite{deng2022stytr2} &0.560 &0.282 \\
        
        (ST) CAST~\cite{zhang2022domain} &0.594 &0.285\\
        
        (ST) QuantArt~\cite{Huang_2023_CVPR} &0.554 &0.289\\ 
        (ST) AesPA-Net~\cite{hong2023aespa} &0.604 &0.284\\
        
        Textual Inversion~\cite{gal2022image} &0.562 &0.278\\
        Custom Diffusion~\cite{kumari2023multi} &0.611 &0.280\\
        DreamBooth~\cite{ruiz2023dreambooth} &0.561 &0.244\\
        StyleDrop~\cite{sohn2023styledrop} &0.602 &0.248\\
        
        \midrule
        \textbf{Ours (\textsc{InstaStyle})} &\textbf{0.655} &\textbf{0.294}\\
        
        \bottomrule
        \end{tabular}
        }}

    \end{minipage}%
    \begin{minipage}[t]{.5\linewidth}
      \centering
        \subcaption{}
        \resizebox{0.9\linewidth}{!}{%
        \setlength{\tabcolsep}{2mm}{
        \begin{tabular}{lccc}
        \toprule
        \textbf{Method} &Preference &Tie &Ours\\
    
        \midrule
        (ST) AdaAttN~\cite{liu2021adaattn} &0.26 &0.13 &\textbf{0.61} \\
        (ST) CCPL~\cite{wu2022ccpl} &0.30 &0.23 &\textbf{0.47}\\
        (ST) StyTr$^{2}$~\cite{deng2022stytr2} &0.27 &0.20 &\textbf{0.53}\\
        (ST) CAST~\cite{zhang2022domain} &0.31 & 0.26 &\textbf{0.43}\\
        (ST) QuantArt~\cite{Huang_2023_CVPR} &0.12 &0.23 &\textbf{0.65}\\ 
        (ST) AesPA-nNet~\cite{hong2023aespa} &0.23 &0.22 &\textbf{0.55}\\
        
        Textual Inversion~\cite{gal2022image}  &0.09 &0.13 &\textbf{0.78}\\
        Custom Diffusion~\cite{kumari2023multi} &0.09 &0.15 &\textbf{0.76} \\
        DreamBooth~\cite{ruiz2023dreambooth}  &0.24 &0.17 &\textbf{0.56}\\
        StyleDrop~\cite{sohn2023styledrop}  &0.27 &0.26 &\textbf{0.47}\\
    
        \bottomrule
        \end{tabular}
        }}
        
    \end{minipage} 
\end{table}

\noindent\textbf{Quantitative results.} We follow evaluation measures in prior works~\cite{gal2022image,sohn2023styledrop}. We utilize 100 objects in CIFAR100~\cite{krizhevsky2009learning} as the target objects.
The generated images are evaluated from two aspects. For style consistency, we calculate the CLIP score~\cite{radford2021learning} between the generated image and the reference image. For content fidelity, CLIP score~\cite{radford2021learning} is calculated between the image and the prompt.

As shown in~\cref{tab:comparison_user} (a), \textsc{InstaStyle} achieves the highest style consistency and content fidelity, indicating that the images generated by \textsc{InstaStyle} exhibit a consistent style with the reference image while retaining its generation capability.
For the comparison methods, Textual Inversion and Custom Diffusion have a lower style consistency, falling short of \textsc{InstaStyle} in style preservation. 
As for DreamBooth and StyleDrop, our approach surpasses them in generating target objects with a higher content score.
We also compare with style transfer methods~\cite{liu2021adaattn,zhang2022domain,wu2022ccpl,deng2022stytr2,Huang_2023_CVPR,hong2023aespa} which utilize content images as input. Notably, our approach diverges by only employing text prompts as content input. This brings challenges in terms of generating high-quality content as text prompts contain less information than content images. Despite this distinction, we achieve an improved content score. Additionally, our approach excels in preserving style details, exhibiting a higher style consistency score.

\noindent\textbf{User study.}
Given the highly subjective nature of stylized generation and the inherent biases in content alignment and style consistency, we conduct a user study following previous works~\cite{deng2022stytr2,sohn2023styledrop}. 
Each question presents participants with a pair of images: one stylized image generated by our approach and another produced by a selected comparison method. To avoid bias, the images are anonymized and randomized for each question.
Participants are tasked with evaluating which result they believed exhibited better stylization effects and content structures. 
To ensure robustness and reliability, we invite 20 participants for the user study. We compare our method with 10 existing methods, including the style transfer methods~\cite{liu2021adaattn,zhang2022domain,wu2022ccpl,deng2022stytr2,Huang_2023_CVPR,Hong_2023_ICCV} and current comparable methods~\cite{gal2022image,kumari2023multi,ruiz2023dreambooth,sohn2023styledrop}. For each method, 30 generated image pairs are randomly selected. Finally, each participant completed 300 rounds of comparisons, resulting in a total of 6,000 votes across all methods.
We count the votes and show the statistical results in~\cref{tab:comparison_user} (b).  
Our approach obtains a higher preference, underscoring its superiority over competitors in both style preservation and content generation.

\begin{figure*}
\begin{center}
\includegraphics[width=\linewidth]{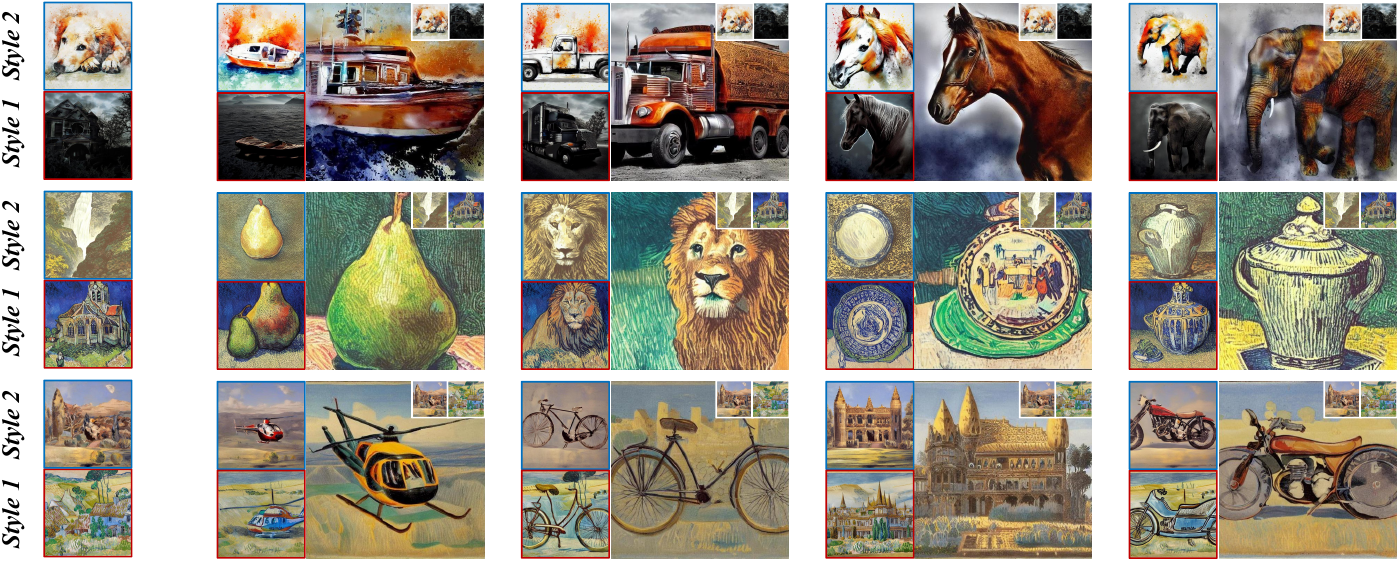}
\end{center}
\caption{\textbf{Visualization of the combination of two styles.} In each case, we show our combination results (the biggest image). 
For better comparison, we also present the stylized generation results of style1 and style2 in the top left (blue box) and bottom left (red box), respectively. Objects for synthesis are \textit{Boat}, \textit{Truck}, \textit{Horse}, \textit{Elephant}, \textit{Pear}, \textit{Lion}, \textit{Plate}, \textit{Pot}, \textit{Helicopter}, \textit{Bicycle}, \textit{Palace}, and \textit{Motorcycle}.}\label{fig:figure_style_combine}
\end{figure*}

\begin{figure}
  \centering
  \includegraphics[width=0.6\textwidth]{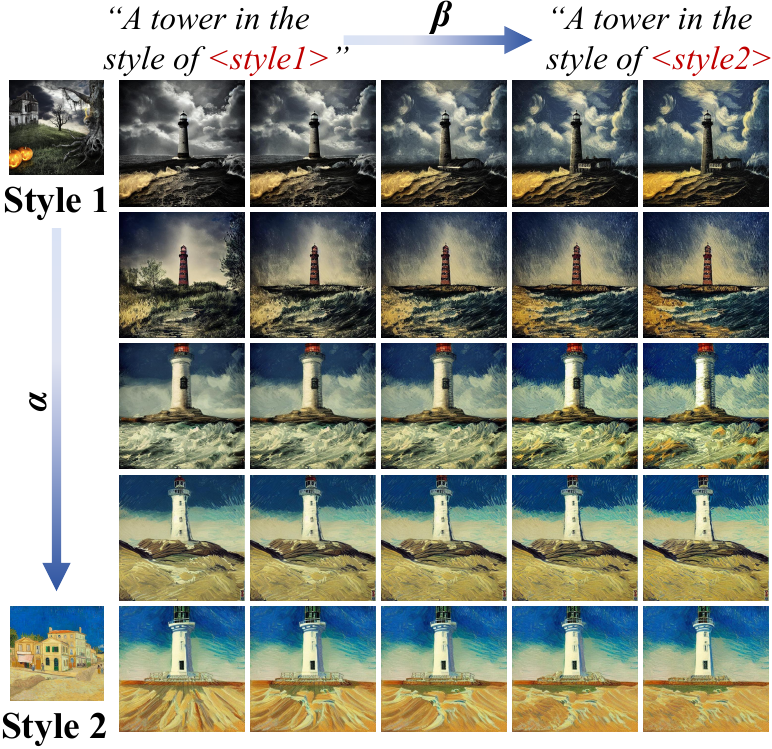}
  \caption{\textbf{Visualization of the combination of two styles.} 
  The style can be controlled via the noise mix ratio and the prompt mix ratio. 
  Our approach enables continuous style combinations, demonstrating its flexibility and diversity.}
  \label{fig:figure_style_combine_scale}
\end{figure}

\subsection{Combination of Two Styles} \label{sec:experiment_sty_sty}
We present style combination results in Fig.~\ref{fig:figure_style_combine}, where we set $\alpha=0.5$ and $\beta=0.5$ to illustrate a more obvious
effect of style combination. Specifically, in each case, we also show the stylized generation results of style 1 (the top left) and style 2 (bottom left), respectively. The biggest image on the right is our combination results, showing the powerful style combination ability of our approach.
More visualizations are shown in.~\cref{fig:sup_figure_stysty_2,fig:sup_figure_stysty_2_2} in Supplementary Material.

As introduced in Sec.~\ref{sec:inference}, both the noise mix ratio $\alpha$ and the prompt mix ratio $\beta$ can affect the style. Fig.~\ref{fig:figure_style_combine_scale} illustrates the impact of the two parameters, where each row shows a different noise mix ratio ($\alpha=0.1, 0.3, 0.5, 0.7, 0.9$) and each column shows a different prompt mix ratio ($\beta=0.1, 0.3, 0.5, 0.7, 0.9$). The style is mainly influenced by the noise and the prompt further improves the style details, showing that our approach is flexible and can generate diverse results.

\subsection{Ablation Study }
\begin{figure}
\begin{center}
\includegraphics[width=\linewidth]{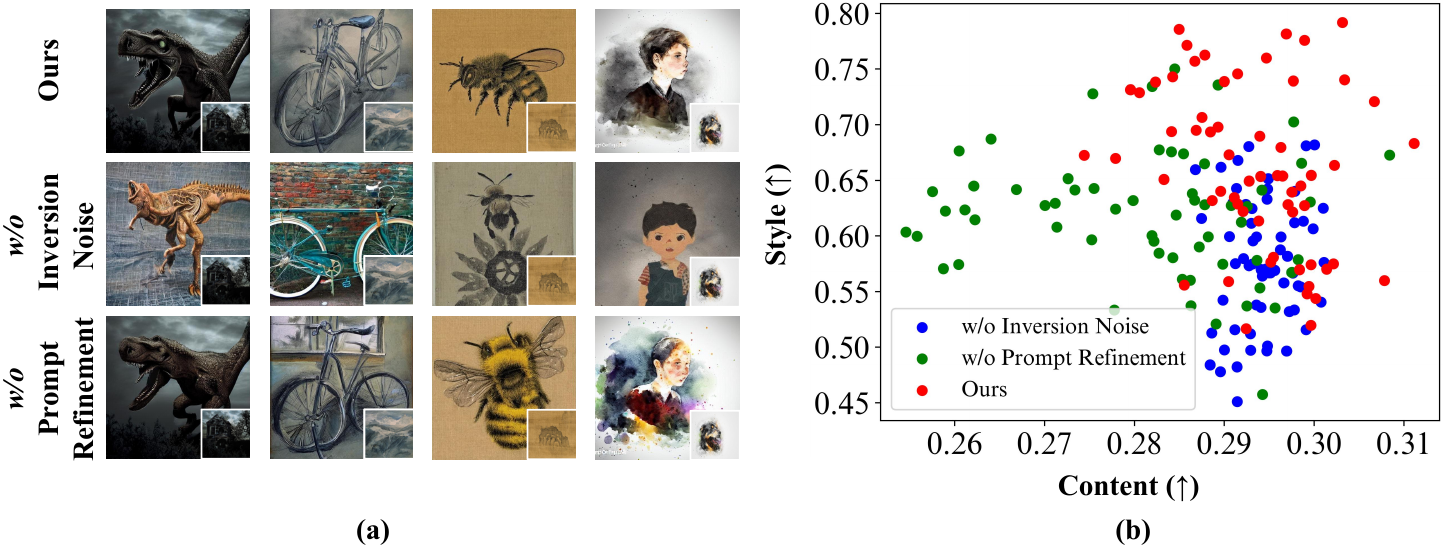}
\end{center}
\caption{\textbf{Ablation study.} (a) Visualization of ablation study. The reference style image is shown in the bottom right corner. Objects for synthesis are \textit{Dinosaur}, \textit{Bicycle}, \textit{Bee}, and \textit{Boy}, respectively. Without the inversion noise, the stylized performance is seriously degraded. Prompt refinement can further improve style details. (b) Quantitative results of ablation study.  Our method lies further along the top right corner, showing better style preservation and content generation capability.}\label{fig:ablation}
\end{figure}

Our approach achieves stylized generation from two perspectives. One is the ``style'' noise obtained via DDIM inversion, which is utilized as the initial noise during inference to provide fine-grained style information. The other is the style token, which is learned via prompt refinement tuning to describe the style precisely. In this section, we conduct the ablation study. The visualization results and quantitative results are shown in  Fig.~\ref{fig:ablation} (a) and Fig.~\ref{fig:ablation} (b), respectively. Specifically, each point in Fig.~\ref{fig:ablation} (b) presents the average style consistency score and content fidelity score of 100 generated images for a reference style. 

\noindent\textbf{Impact of inversion noise.} 
A prominent advantage of our \textsc{InstaStyle} is that we utilize the inversion noise as the initial image during inference time to preserve the fine-grained style information. Thereby, we first conduct an ablation study by replacing the inversion ``style'' noise with random noise.
% The visualization results in Fig.~\ref{fig:ablation} (third row
As shown in Fig.~\ref{fig:ablation} (a) (second row), without the inversion noise, the stylized performance is seriously degraded.
The quantitative results in Fig.~\ref{fig:ablation} (b) further illustrate that ablating the ``style'' noise (blue dots) will result in a lower style consistency score, harming the style of the generated image.

\noindent\textbf{Impact of prompt refinement tuning.}
We ablate the prompt refinement stage to show the necessity
of learning a style token to avoid ambiguity and bias in human-writing textual style tokens. Fig.~\ref{fig:ablation} (a) (third row) reveals that our learnable style token can better describe the style without harming the generation ability. The quantitative results in Fig.~\ref{fig:ablation} (b) show that ablating the prompt refinement (green dots) will result in a lower content fidelity score and style score, further proving the importance of the learnable style token.

%% file: sec/5_conclusion.tex
\section{Conclusion}
\label{sec:conclusion}
Our research solves the task of stylized generation by exploring the fine-grained style information in the reference image. Through extensive analysis and empirical investigation, we find that the initial noise with a non-zero signal-to-noise ratio in the diffusion model leans toward a particular style. Therefore, DDIM inversion is introduced to obtain an initial noise containing fine-grained style information. Furthermore, we highlight the challenge caused by the internal bias in textual style tokens and propose to learn a style token adaptively to better describe the style without bias. Our approach produces satisfactory stylized generation results and supports the combination of two styles. 
Extensive qualitative and quantitative comparisons demonstrate the superiority of our method.

\noindent{\textbf{Acknowledgement}}
This research is sponsored by National Natural Science Foundation of China (Grant No. 62306041, U21B2045, 62176025), Beijing Nova Program (Grant No. Z211100002121106, 20230484488, 20230484276), Youth Innovation Promotion Association CAS (Grant No.2022132), and Beijing Municipal Science \& Technology Commission (Z231100007423015).

%% file: sec/X_suppl.tex
\clearpage
\setcounter{page}{1}

\appendix
{\centering

\textbf{\textsc{InstaStyle}: Inversion Noise of a Stylized Image is Secretly a Style Adviser}\\
\vspace{0.5em}Supplementary Material \\
\vspace{1.0em}
}

In this supplementary material, we first present additional preliminaries for the diffusion model in Sec.~\ref{sup:addition_preliminaries}. Subsequently, the derivation of the signal-to-noise ratio (SNR) of the inversion noise is shown in Sec.~\ref{sup:snr}. 
In Sec.~\ref{sup: more_experiments}, we show more experiment details and experiment results, including implementation details (Sec.~\ref{sup: implementation_detail}), more qualitative results (Sec.~\ref{sup:more_qualitative_results}) and more analysis (Sec.~\ref{sup:analysis}).
% \vspace{-50mm}
\begin{figure*}
    \centering
    \begin{subfigure}{\linewidth}
        \centering
        \includegraphics[width=\linewidth]{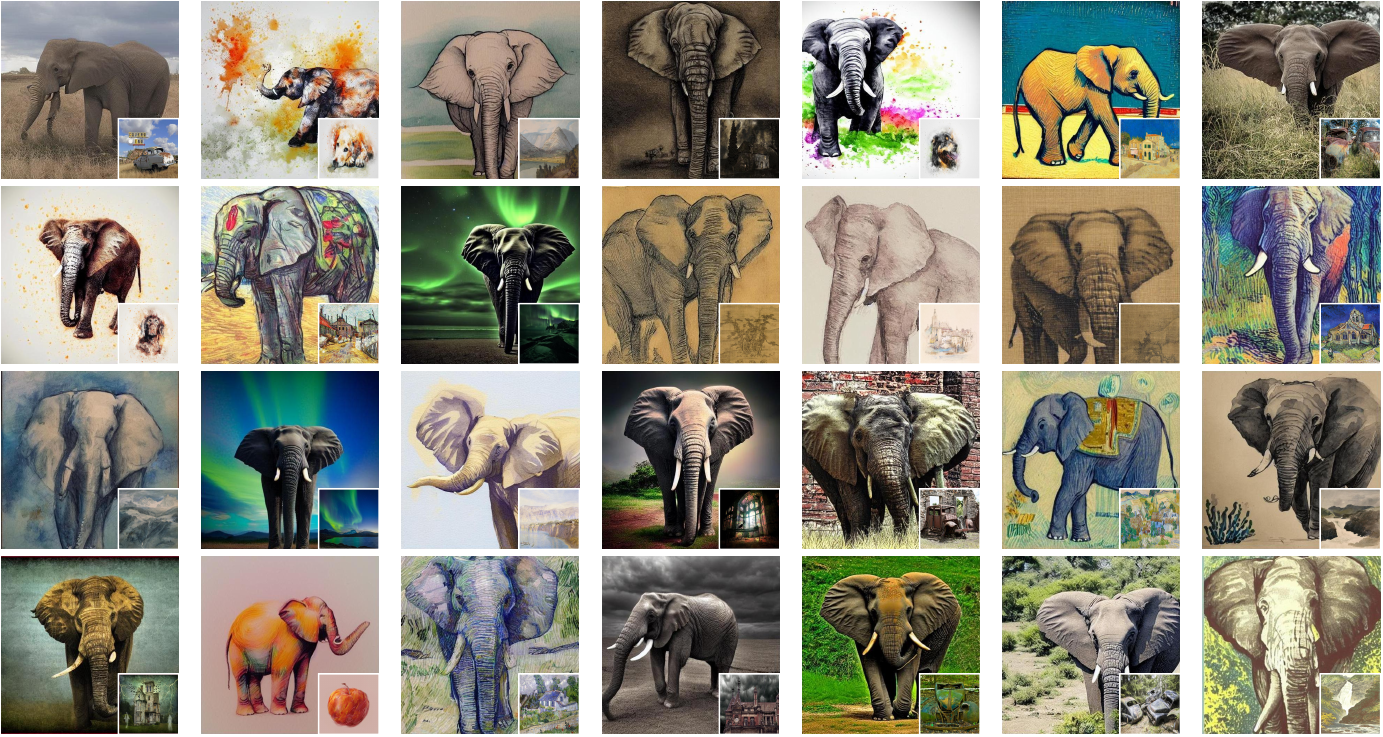}
        \subcaption{The object for synthesis is \textit{Elephant}.}
    \end{subfigure}
    
    % \vspace{5mm} % 控制两张图片之间的垂直间距
    
    \begin{subfigure}{\linewidth}
        \centering
        \includegraphics[width=\linewidth]{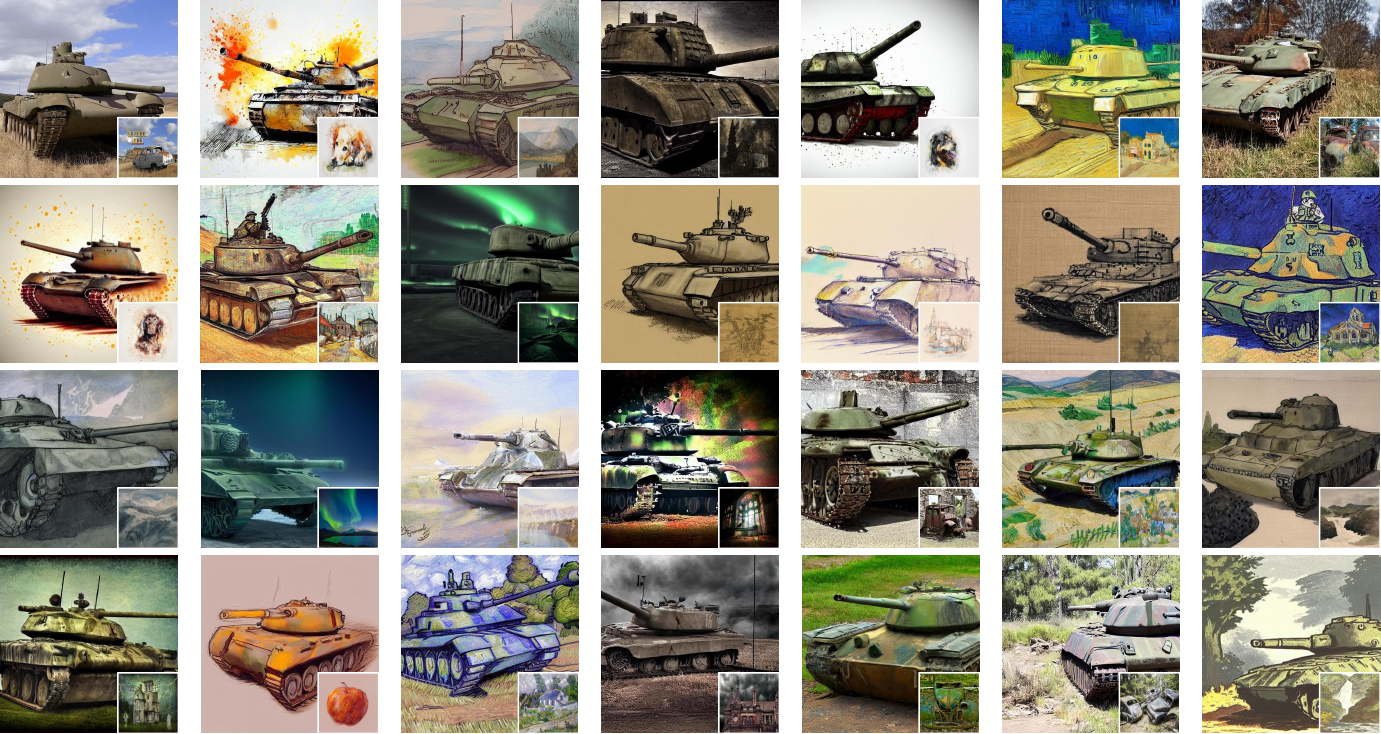}
        \subcaption{The object for synthesis is \textit{Tank}.}
    \end{subfigure}
    
    \caption{\textbf{Qualitative results on various reference styles.}}
    \label{fig:sup_figure_elephant_tank}
     \vspace{-15mm}
\end{figure*}

\begin{figure*}
  \centering

  \begin{subfigure}{1\textwidth}
    \centering
    \includegraphics[width=0.95\linewidth]{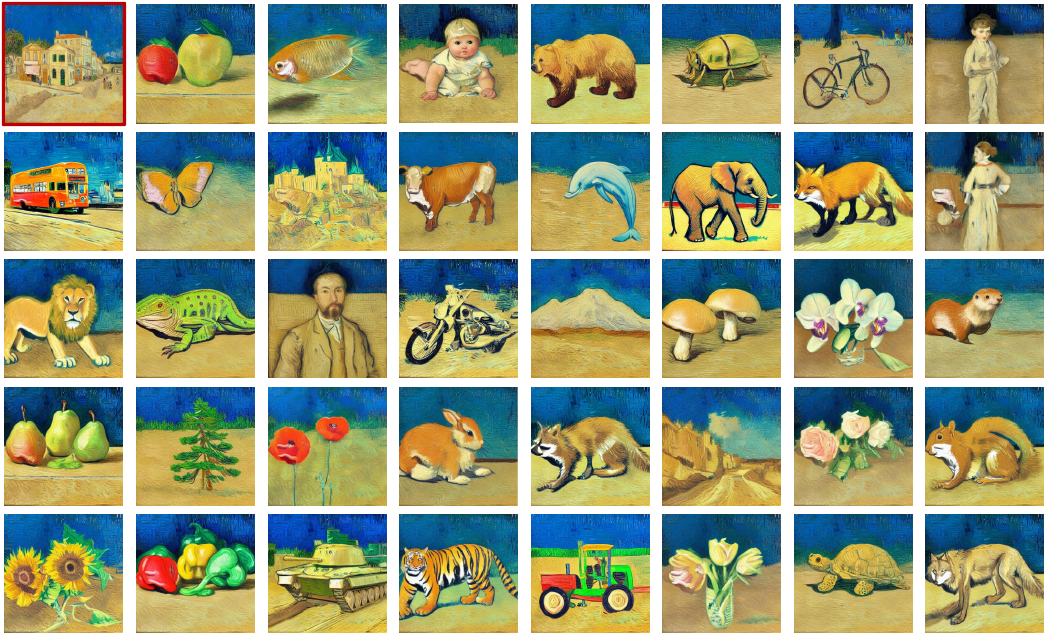}
    \subcaption*{(a)}
  \end{subfigure}
% ~\\
  \begin{subfigure}{1\textwidth}
    \centering
    \includegraphics[width=0.95\linewidth]{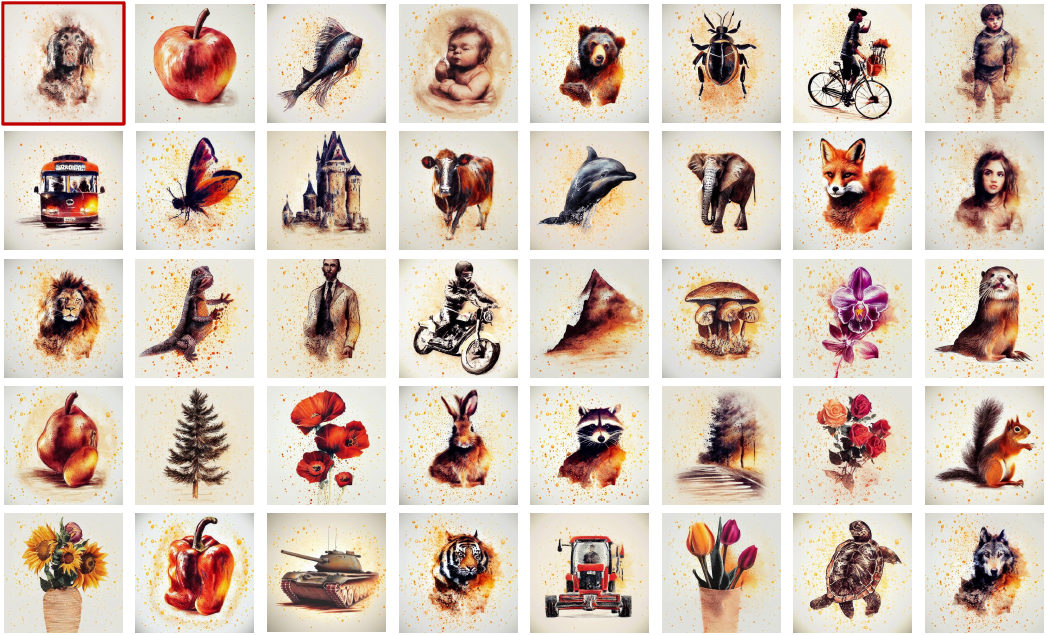}
    \subcaption*{(b)}
  \end{subfigure}

\caption{\textbf{Qualitative results on various objects.}  The image on the top left is the reference style image. Objects for synthesis are {\textit{Apple}}, {\textit{Aqariumfish}}, {\textit{Baby}}, {\textit{Bear}}, {\textit{Beetle}}, {\textit{Bicycle}}, {\textit{Boy}}, {\textit{Bus}}, {\textit{Butterfly}}, {\textit{Castle}}, {\textit{Cattle}}, {\textit{Dolphin}}, {\textit{Elephant}}, {\textit{Fox}}, {\textit{Girl}}, {\textit{Lion}}, {\textit{Lizard}}, {\textit{Man}}, {\textit{Motorcycle}}, {\textit{Mountain}}, {\textit{Mushrooms}}, {\textit{Orchids}}, {\textit{Otter}}, {\textit{Pears}},{\textit{Pine}}, {\textit{Poppies}}, {\textit{Rabbit}}, {\textit{Raccoon}}, {\textit{Road}}, {\textit{Roses}}, {\textit{Squirrel}}, {\textit{Sunflowers}}, {\textit{Sweetpeppers}}, {\textit{Tank}}, {\textit{Tiger}}, {\textit{Tractor}}, {\textit{Tulips}}, {\textit{Turtle}}, and {\textit{Wolf}}.}
\label{fig:sup_figure_vangogh_and_water}
\end{figure*}

\begin{figure*}
    \centering
    \begin{subfigure}{\linewidth}
        \centering
        \includegraphics[width=\linewidth]{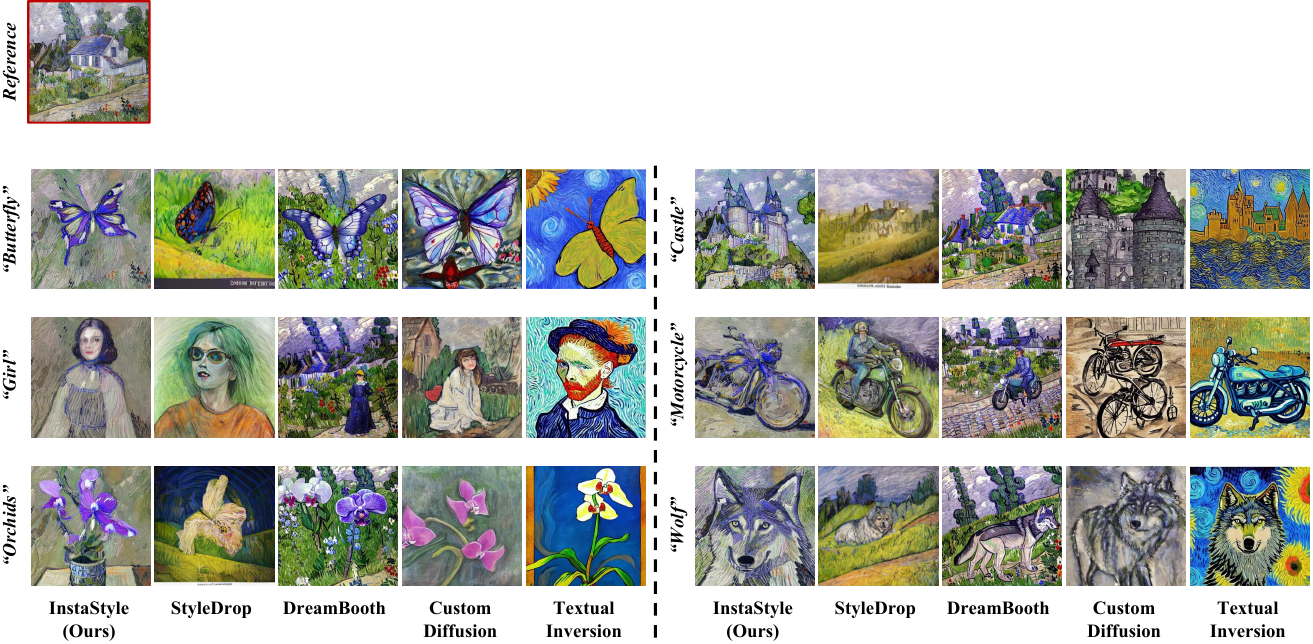}
        \subcaption*{(a)}
    \end{subfigure}
    
    % \vspace{5mm} % 控制两张图片之间的垂直间距
    
    \begin{subfigure}{\linewidth}
        \centering
        \includegraphics[width=\linewidth]{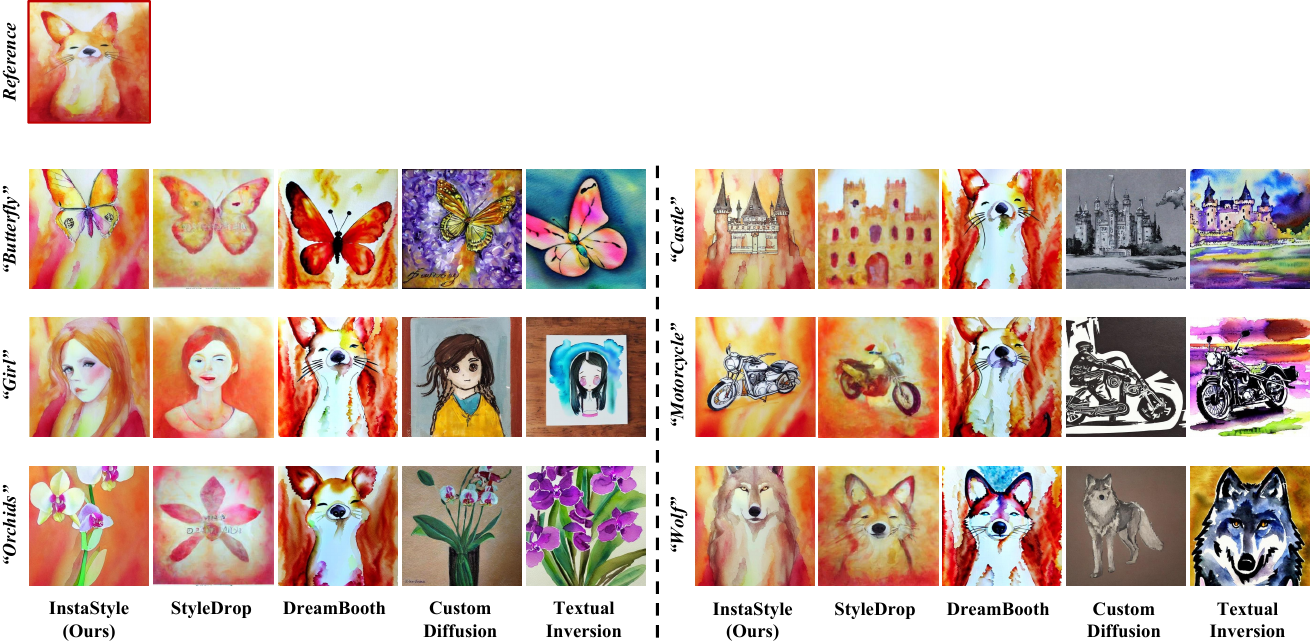}
        \subcaption*{(b)}
    \end{subfigure}
    
    \caption{\textbf{More comparison results.} Objects for synthesis are \textit{Butterfly}, \textit{Castle}, \textit{Girl}, \textit{Motorcycle}, \textit{Orchids}, and \textit{Wolf}. Our \textsc{InstaStyle} exhibits better performance in both style preservation and content generation.}
    \label{fig:sup_figure_compare_a_b}
     \vspace{-15mm}
\end{figure*}

\begin{figure*}
    \centering
    \begin{subfigure}{\linewidth}
        \centering
        \includegraphics[width=\linewidth]{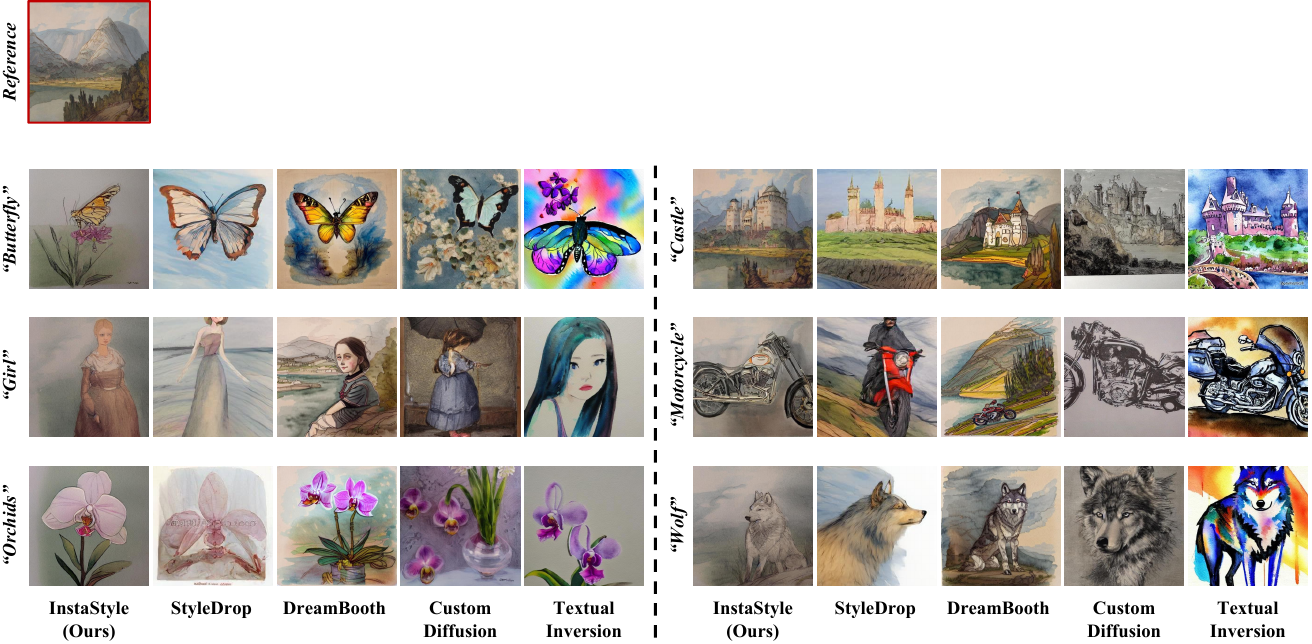}
        \subcaption*{(a)}
    \end{subfigure}
    
    % \vspace{10mm} % 控制两张图片之间的垂直间距
    
    \begin{subfigure}{\linewidth}
        \centering
        \includegraphics[width=\linewidth]{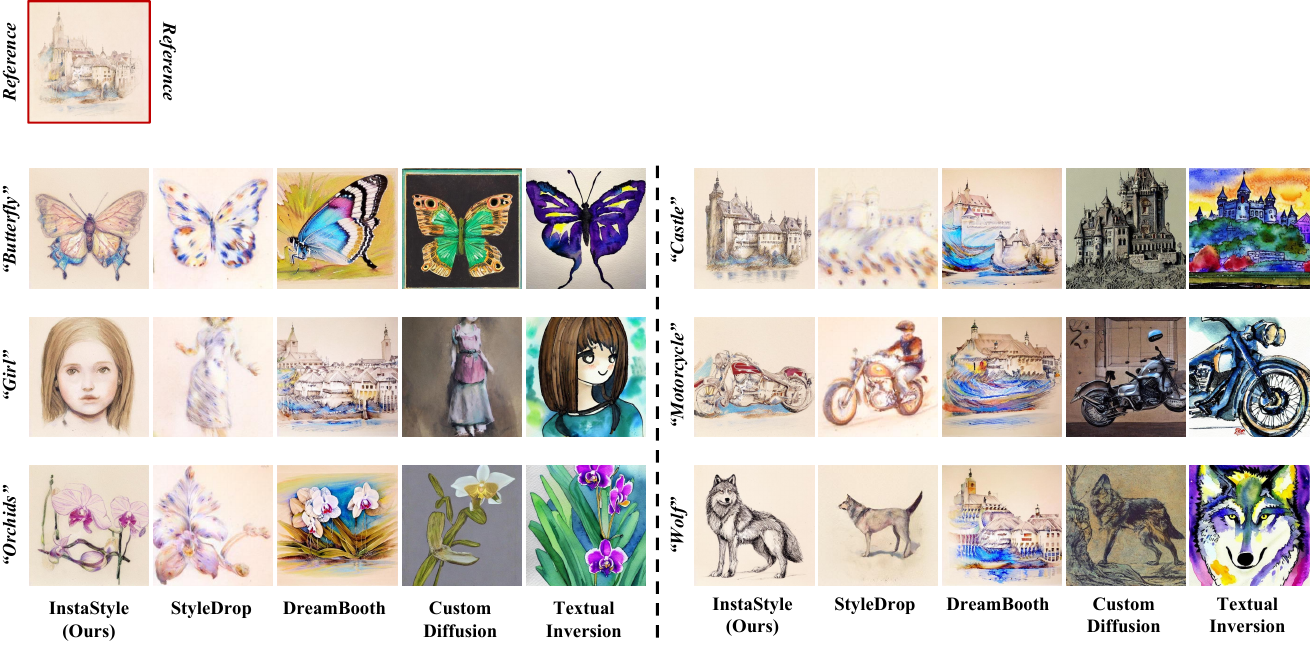}
        \subcaption*{(b)}
    \end{subfigure}
    
    \caption{\textbf{More comparison results.} Objects for synthesis are \textit{Butterfly}, \textit{Castle}, \textit{Girl}, \textit{Motorcycle}, \textit{Orchids}, and \textit{Wolf}. Our \textsc{InstaStyle} exhibits better performance in both style preservation and content generation.}
    \label{fig:sup_figure_compare_c_d}
     \vspace{-15mm}
\end{figure*}

\begin{figure*}[ht]
\begin{center}
\includegraphics[width=0.92\linewidth]{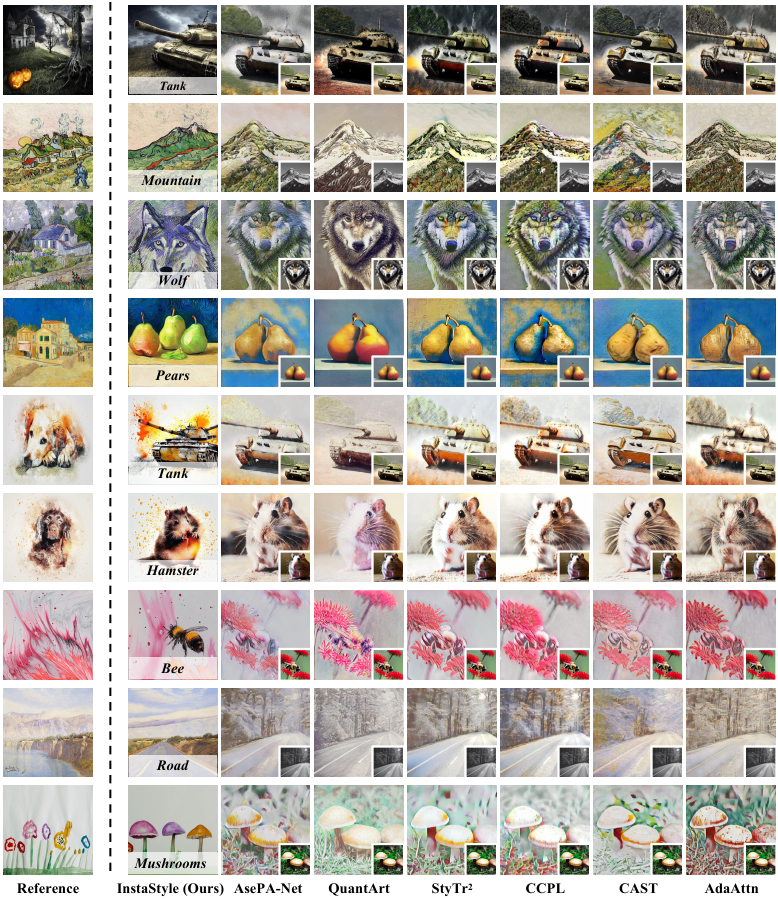}
\end{center}
\vspace{-3mm}
\caption{\textbf{More comparison results with style transfer methods.} }
\label{fig:sup_figure_stytransfer}
\end{figure*}

\begin{figure*}[ht]
\begin{center}
\includegraphics[width=0.92\linewidth]{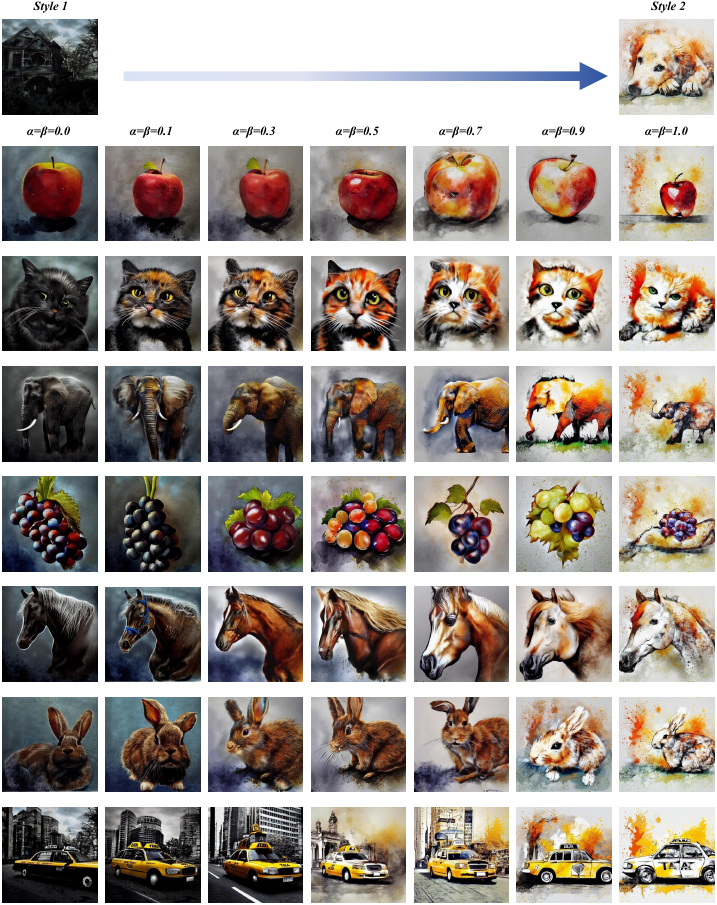}
\end{center}
\vspace{-3mm}
\caption{\textbf{More style combination results.} Our approach supports adjusting the degree of two styles during combination and can generate various target objects,  demonstrating the flexibility and universality of our approach. The noise mix ratio and the prompt mix ratio are set to be equal, \textit{i.e.}, 0, 0.1, 0.3, 0.5, 0.7, 0.9, and 1 from left to right. Objects for synthesis are \textit{Apple}, \textit{Cat}, \textit{Elephant}, \textit{Grape}, \textit{Horse}, \textit{Rabbit}, and \textit{Taxi}.}
\label{fig:sup_figure_stysty_2}
\end{figure*}

\begin{figure*}[ht]
\begin{center}
\includegraphics[width=0.92\linewidth]{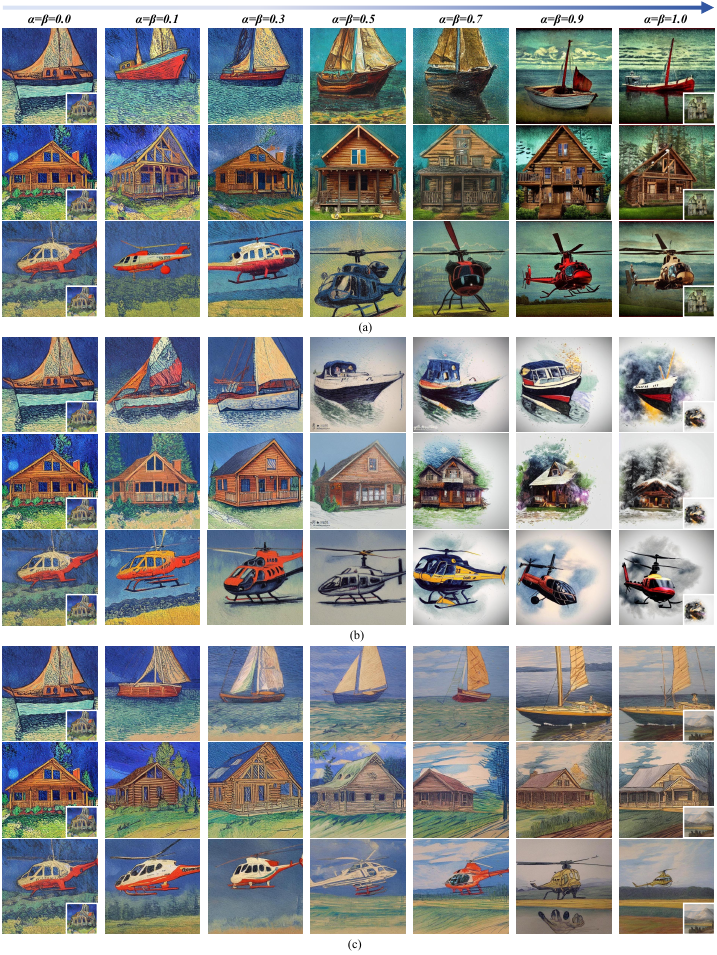}
\end{center}
\vspace{-6mm}
\caption{\textbf{More style combination results.} We present style combination results of a Van Gogh painting with different reference style images, illustrating the creative ability of our method,. The noise mix ratio and the prompt mix ratio are set to be equal, which are 0, 0.1, 0.3, 0.5, 0.7, 0.9, and 1 from left to right. The reference images are shown in the bottom right corner of the left most image and the right most image, respectively. Objects for synthesis are \textit{Boat}, \textit{Cabin}, and \textit{Helicopter}.} 
\label{fig:sup_figure_stysty_2_2}
\end{figure*}

\clearpage

\section{Additional Preliminaries} \label{sup:addition_preliminaries}
Diffusion Denoising Probabilistic Model~\cite{ho2020denoising} is a generative model that aim to approximate the data distribution $q(z_0)$ with a distribution $p_{\theta}(z_0)$. Generating a new image is equivalent to sampling from the data distribution $q(z_0)$. 
In practice, we utilize a backward process  $q(z_{t-1}|z_t)$ to iteratively denoise from a Gaussian noise $z_T$.
Since the backward process $q(z_{t-1}|z_t)$ depends the unknown data distribution $q(z_0)$, a parameterized Gaussian transition network $p_\theta(z_{t-1}|z_t):=\mathcal{N}(z_{t-1};\mu_{\theta}(z_{t},t),\Sigma_{\theta}(z_{t},t))$ is introduced to approximate $q(z_{t-1}|z_t)$. The $\mu_{\theta}(z_{t},t)$ can be reparameterized as:
\begin{equation}
    \mu_\theta(z_t,t)=\frac1{\sqrt{\alpha_t}}\left(z_t-\frac{\beta_t}{\sqrt{1-\alpha_t}}\varepsilon_\theta(z_t,t)\right),
\end{equation}
% Diffusion models~\cite{ho2020denoising,sohl2015deep} contain a forward process and a backward process. The forward process adds noise to the data according to a predetermined, non-learned variance schedule $\beta _{1},...,\beta _{T}$, which is defined as a Gaussian transition:
% \begin{equation}
%     q(z_t|z_{t-1}):=\mathcal{N}(z_t;\sqrt{1-\beta_t}z_{t-1},\beta_t\mathbf{I}).
% \end{equation}
% An important property of the forward process is that we can obtain the latent variable $z_t$ directly based on $z_0$:
% \begin{equation}
%     \label{eq:qt_q0}
%     z_t=\sqrt{\alpha_t}z_0+\sqrt{1-\alpha_t}\varepsilon,
% \end{equation}
% where $\varepsilon\sim\mathcal{N}(\mathbf{0},\mathbf{I})$, $\alpha_t:=\prod_{i=1}^t(1-\beta_i)$.
% Diffusion models restore the information by learning the backward process:
% \begin{equation}
%     p_\theta(z_{t-1}|z_t):=\mathcal{N}(z_{t-1};\mu_{\theta}(z_{t},t),\Sigma_{\theta}(z_{t},t)).
% \end{equation}
% Details are shown in Sec.~\ref{sup:addition_preliminaries} in Supplementary Materials.
% where $\varepsilon_\theta(z_t,t)$ is the predicted noise added to $z_0$. 
where $\beta _{t}$ is predetermined variance schedule, $\alpha_t:=\prod_{i=1}^t(1-\beta_i)$. $z_t$ is obtained by adding an artificial noise $\varepsilon\sim\mathcal{N}(\mathbf{0},\mathbf{I})$ to $z_0$, \ie, $z_t=\sqrt{\alpha_t}z_0+\sqrt{1-\alpha_t}\varepsilon$. $\varepsilon_\theta(z_t,t)$ is the learnable network that predicts the artificial noise.
Once we have trained $\varepsilon_\theta(z_t,t)$, we can iteratively denoise from a Gaussian noise $z_T$ as follows:
\begin{equation}
    z_{t-1}=\mu_\theta(z_t,t)+\sigma_tz,z\sim N(0,I).
\end{equation}
The $\sigma_t$ of each sample stage has different settings in different denoising approaches. For example, in DDIMs~\cite{song2020denoising}, the denoising process is made to be deterministic by setting $\sigma_t=0$. 

The training process is to learn $\varepsilon_\theta(z_t,t)$ which predicts the artificial noise added to the current image:
\begin{equation}
        \min_{\theta}E_{z_0,\varepsilon\sim N(0,I),t\sim\mathrm{Uniform}(1,T)}\left\|\varepsilon-\varepsilon_\theta(z_t,t)\right\|_2^2.
    \label{eq:dm_loss}
\end{equation}

It is worth noting that the fundamental diffusion model~\cite{ho2020denoising} is an unconditional model, while the Stable Diffusion~\cite{rombach2022high} utilized in our framework is a conditional model. Therefore, compared to the optimization target in~\cref{eq:sd_loss}, ~\cref{eq:dm_loss} does not have the conditional term $\mathcal{C}$.

\section{SNR of Inversion Noise}\label{sup:snr}
\label{sec:SNR_DDIM_Inversion}
As $\varepsilon_{\theta}(z_{t},t,\mathcal{C})$ is trained to approximate the artificial noise $\varepsilon\sim\mathcal{N}(\mathbf{0},\mathbf{I})$, we can assume that $\varepsilon_{\theta}(z_{t},t,\mathcal{C})\sim\mathcal{N}(\mathbf{0},\mathbf{I})$. For simplicity, we denote $\varepsilon_{\theta}(z_{t},t,\mathcal{C})$ as $\varepsilon _{t}$. 
According to~\cref{eq:ddim_inverion}, $z_{t+1}$ can be approximately obtained in closed form:

\begin{equation}
\resizebox{0.9\textwidth}{!}{
$\
\begin{aligned}
z_{t+1} 
&=\sqrt{\frac{\alpha _{t+1}}{\alpha _{t}}}z_{t}+\left( \sqrt{\frac{1}{\alpha _{t+1}}-1}-\sqrt{\frac{1}{\alpha _{t}}-1} \right) \cdot \varepsilon _{t} \\
&=\sqrt{\frac{\alpha _{t+1}}{\alpha _{t}}}\left( \sqrt{\frac{\alpha _{t}}{\alpha _{t-1}}}z_{t-1}+\left( \sqrt{\frac{1}{\alpha _{t}}-1}-\sqrt{\frac{1}{\alpha _{t-1}}-1} \right) \cdot \varepsilon _{t-1} \right) +\left( \sqrt{\frac{1}{\alpha _{t+1}}-1}-\sqrt{\frac{1}{\alpha _{t}}-1} \right) \cdot \varepsilon _t \\
&=\sqrt{\frac{\alpha _{t}}{\alpha _{t-1}}}z_{t-1}+\sqrt{\frac{\alpha _{t+1}}{\alpha _t}}\left( \sqrt{\frac{1}{\alpha _{t}}-1}-\sqrt{\frac{1}{\alpha _{t-1}}-1} \right) \cdot \varepsilon _{t-1}+\left( \sqrt{\frac{1}{\alpha _{t+1}}-1}-\sqrt{\frac{1}{\alpha _{t}}-1} \right) \cdot \varepsilon _{t} \\
&=\sqrt{\frac{\alpha _{t}}{\alpha _{t-1}}}z_{t-1}+\sqrt{\left( \sqrt{\frac{\alpha _{t+1}}{\alpha _{t}}}\left( \sqrt{\frac{1}{\alpha _t}-1}-\sqrt{\frac{1}{\alpha _{t-1}}-1} \right) \right) ^2+\left( \sqrt{\frac{1}{\alpha _{t+1}}-1}-\sqrt{\frac{1}{\alpha _{t}}-1} \right) ^2} \cdot \bar{\varepsilon}_{t-1} \\
&=\sqrt{\frac{\alpha _{t}}{\alpha _{t-1}}}z_{t-1}+\sqrt{\frac{\alpha _{t+1}}{\alpha _{t}}\left( \sqrt{\frac{1}{\alpha _{t}}-1}-\sqrt{\frac{1}{\alpha _{t-1}}-1} \right) ^2+\frac{\alpha _{t+1}}{\alpha _{t+1}}\left( \sqrt{\frac{1}{\alpha _{t+1}}-1}-\sqrt{\frac{1}{\alpha _{t}}-1} \right) ^2} \cdot \bar{\varepsilon}_{t-1} \\
&=... \\
&=\sqrt{\frac{\alpha _{t}}{\alpha _0}}z_0+ \sqrt{\sum_{i=0}^t{\frac{\alpha _{t+1}}{\alpha _{i+1}}\left( \sqrt{\frac{1}{\alpha _{i+1}}-1}-\sqrt{\frac{1}{\alpha _i}-1} \right) ^2}} \cdot \bar{\varepsilon}_{0},
\end{aligned}
$
}
\end{equation}

where $\varepsilon _{t}, \varepsilon _{t-1},\cdots \varepsilon _{1}\sim\mathcal{N}(\mathbf{0},\mathbf{I})$. $\bar{\varepsilon}_{t-1}$ merges two Gaussions $\varepsilon _{t}$ and $\varepsilon _{t-1}$. $\bar{\varepsilon}_{t-1},\bar{\varepsilon}_{t-2},\bar{\varepsilon}_{0}\sim\mathcal{N}(\mathbf{0},\mathbf{I})$. 
Following~\cite{rombach2022high,lin2023common}, the signal-to-noise ratio (SNR) can be calculated as:
\begin{equation}
\resizebox{0.6\textwidth}{!}{
$\begin{aligned}
    \mathrm{SNR}(t):&=
\frac{\left( \sqrt{\frac{\alpha _t}{\alpha _0}} \right) ^2}{\left( \sqrt{\sum_{i=0}^t{\frac{\alpha _{t+1}}{\alpha _{i+1}}\left( \sqrt{\frac{1}{\alpha _{i+1}}-1}-\sqrt{\frac{1}{\alpha _i}-1} \right) ^2}} \right) ^2}\\
    &=\frac{1}{\sum_{i=0}^t{\frac{\alpha _{0}}{\alpha _{i+1}}\left( \sqrt{\frac{1}{\alpha _{i+1}}-1}-\sqrt{\frac{1}{\alpha _i}-1} \right) ^2}}
\end{aligned}$
.
}
\end{equation}

\begin{table}
    \centering
    \caption{Image sources for experiments.}
    \label{tab:supp_image_source}
    \resizebox{\linewidth}{!}{%
    \begin{tabular}{c|c|l}
    \toprule
        \textbf{Style} &\textbf{Object} & \textbf{Link} \\ \midrule
        \multirow{8}{*}{Van Gogh painting}&House & \url{https://www.rawpixel.com/image/3868934/illustration-image-art-vincent-van-gogh-person} \\
&House & \url{https://www.rawpixel.com/image/3865273/illustration-image-art-vincent-van-gogh-house} \\
&House & \url{https://www.rawpixel.com/image/3866294/illustration-image-art-vincent-van-gogh-house} \\
&House & \url{https://www.rawpixel.com/image/3868302/illustration-image-art-vincent-van-gogh-house} \\
&House & \url{https://www.rawpixel.com/image/3864574/illustration-image-art-vincent-van-gogh} \\
&House &\url{https://www.rawpixel.com/image/3864611/illustration-image-art-vincent-van-gogh-house} \\
&House & \url{https://www.rawpixel.com/image/537424/free-illustration-image-van-gogh-factory} \\
&House & \url{https://www.rawpixel.com/image/537422/free-illustration-image-van-gogh-cottage} \\\midrule
\multirow{9}{*}{Haunted}&House & \url{https://search-production.openverse.engineering/image/17662d45-aa50-474e-83b6-bd170eda9bd9} \\
&House & \url{https://search-production.openverse.engineering/image/f010aca4-14d0-4464-a706-b66ed7fb8569} \\
&House & \url{https://search-production.openverse.engineering/image/ce8bd8d7-a509-4ece-a9e2-eb8581d0fb00} \\
&House & \url{https://search-production.openverse.engineering/image/2f1a1eee-c033-4781-bad0-3e12212a2361} \\
&House & \url{https://search-production.openverse.engineering/image/89e86034-fe58-4f32-8259-9dbc4bac8ebc} \\
&House & \url{https://search-production.openverse.engineering/image/089c1cb0-f070-4d1f-9fc5-a82a7939d411} \\
&House & \url{https://www.rawpixel.com/image/5906049/photo-image-public-domain-house-halloween} \\
&House & \url{https://www.rawpixel.com/image/5964852/free-public-domain-cc0-photo} \\
&House & \url{https://www.rawpixel.com/image/6051791/free-public-domain-cc0-photo} \\\midrule
\multirow{9}{*}{Abandoned}&Car & \url{https://search-production.openverse.engineering/image/9e826ffa-bac1-4892-b78a-04eded1cefcf} \\
&Car & \url{https://search-production.openverse.engineering/image/545bb05c-cafc-42fc-89cc-34b0e51593a2} \\
&Car & \url{https://search-production.openverse.engineering/image/b1e4bc4e-982e-4792-a513-bdea0a5f72cb} \\
&Car & \url{https://search-production.openverse.engineering/image/b103ce3b-c74c-41c4-97e0-8c3a7563bd59} \\
&Car & \url{https://www.rawpixel.com/image/5941829/free-public-domain-cc0-photo} \\
&Car & \url{https://www.rawpixel.com/image/4026101/oldsmobile-route-66} \\
&Car & \url{https://search-production.openverse.engineering/image/fbb2922a-a03c-44e7-9571-6901bba27957} \\
&Car & \url{https://search-production.openverse.engineering/image/f579052d-d3c8-4a59-8217-97b9f4b6ae55} \\
&Car & \url{https://search-production.openverse.engineering/image/f8db9513-f0a6-40e0-9e23-e6e34b5e04f3} \\\midrule
\multirow{3}{*}{Watercolor}&Dog & \url{https://pixy.org/6458158/} \\
&Dog & \url{https://pixy.org/5792070/} \\
&Dog & \url{https://pixy.org/6379346/} \\
&Dog & \url{https://pixy.org/5782945/} \\\midrule
\multirow{5}{*}{Chinese inkpainting}&Mountain & \url{https://search-production.openverse.engineering/image/382e0452-14a5-4895-aff2-917dfec8f40c} \\
&Mountain & \url{https://search-production.openverse.engineering/image/22a3f474-128d-4a69-8d4e-98f904a77170} \\
&Mountain & \url{https://search-production.openverse.engineering/image/3aa5eee8-3a49-4e8e-9df9-f6ca2e0dc01c} \\
&Mountain & \url{https://search-production.openverse.engineering/image/293879fc-9f71-410f-b96b-161493e14377} \\
&Mountain & \url{https://search-production.openverse.engineering/image/04edb78b-b848-430b-9166-c61efab08a3c} \\\midrule
\multirow{8}{*}{Aurora}&Mountain & \url{https://unsplash.com/photos/CgoRzWX4CDg} \\
&Mountain & \url{https://unsplash.com/photos/U_diPCXCBxU} \\
&Mountain & \url{https://unsplash.com/photos/-OkHUsepnzw} \\
&Mountain & \url{https://unsplash.com/photos/pDeagUyN-Pk} \\
&Mountain & \url{https://unsplash.com/photos/ZJDMls6ppY8} \\
&Mountain & \url{https://unsplash.com/photos/Hn8N4I4eHA0} \\
&Mountain & \url{https://unsplash.com/photos/fpaSXDuoHkc} \\
&Mountain & \url{https://unsplash.com/photos/w1yDuFs-kGY} \\\midrule
\multirow{3}{*}{Inkpainting}&Mountain & \url{https://unsplash.com/photos/6fv0MEf3FUE} \\
&Mountain & \url{https://unsplash.com/photos/NspHfyZnMBE} \\
&Mountain & \url{https://unsplash.com/photos/Vc8GBqapdfs} \\\midrule
\multirow{1}{*}{Oilpainting}&Apple & \url{https://unsplash.com/photos/LQTdG9SJpyA} \\\midrule
\multirow{9}{*}{Watercolor}&Dog & \url{https://unsplash.com/photos/KRztl5I6xac} \\
&Mountain & \url{https://unsplash.com/photos/0pJPixfGfVo} \\
&Flowers & \url{https://unsplash.com/photos/YIfFVwDcgu8} \\
&House & \url{https://unsplash.com/photos/9dnNnTrHxmI} \\
&Feather & \url{https://unsplash.com/photos/Tyg0rVhOTrE} \\
&Fox & \url{https://unsplash.com/photos/8D-0K6JUAEE}\\
&Dress & \url{https://unsplash.com/photos/6NSVToSYwV0} \\
&Flowers & \url{https://unsplash.com/photos/-KfLa4I4eTo}\\
&House & \url{https://unsplash.com/photos/X2QwsspYk_0} \\
&Mountain & \url{https://unsplash.com/photos/wvD0zZnRbcw} \\
&Mountain & \url{https://unsplash.com/photos/-IAS_N85adA}\\
&Mountain & \url{https://unsplash.com/photos/TAZga9MibgA}\\
&Flowers & \url{https://unsplash.com/photos/6dY9cFY-qTo}\\

    \bottomrule
    \end{tabular}
    }
\end{table}

\begin{table}[h]
\centering
\caption{Objects for generation in the quantitative experiment.}
\label{tab:supp_obj_stage2}
\resizebox{0.6\linewidth}{!}{%
\begin{tabular}{l|c}
\toprule
Superclass & Objects  \\
\midrule
aquatic & mammals beaver, dolphin, otter, seal, whale \\
large natural outdoor scenes & cloud, forest, mountain, plain, sea \\
fish & aquarium fish, flatfish, ray, shark, trout \\ 
large omnivores and herbivores & camel, cattle, chimpanzee, elephant, kangaroo \\
flowers & orchids, poppies, roses, sunflowers, tulips \\ 
medium-sized mammals & fox, porcupine, possum, raccoon, skunk \\
food containers & bottles, bowls, cans, cups, plates \\
non-insect invertebrates & crab, lobster, snail, spider, worm \\
fruit and vegetables & apples, mushrooms, oranges, pears, sweet peppers\\
people & baby, boy, girl, man, woman \\
household electrical devices & clock, computer keyboard, lamp, telephone, television \\
reptiles & crocodile, dinosaur, lizard, snake, turtle \\
household furniture & bed, chair, couch, table, wardrobe \\
small mammals & hamster, mouse, rabbit, shrew, squirrel \\
insects & bee, beetle, butterfly, caterpillar, cockroach \\
trees & maple, oak, palm, pine, willow \\
large carnivores & bear, leopard, lion, tiger, wolf \\
vehicles 1 & bicycle, bus, motorcycle, pickup truck, train \\
large man-made outdoor things & bridge, castle, house, road, skyscraper \\
vehicles 2 & lawn-mower, rocket, streetcar, tank, tractor \\
\bottomrule
\end{tabular}}
\end{table}

\section{Experiments}\label{sup: more_experiments}
\subsection{Implementation Detail}\label{sup: implementation_detail}
\noindent\textbf{Datasets.}
The reference image sources for experiments are presented in~\cref{tab:supp_image_source}. We also label the name of the style and object in the reference image, which is utilized in the first stage to transform the reference image into a ``style'' noise. Besides, the name of the style is also used to initialize the learnable style token.

\noindent\textbf{Objects for generation.}
In the first stage, we generate 15 objects for each reference image which is utilized to fine-tune the learnable style token in the second stage. Specifically, these objects are \textit{cat}, \textit{lighthouse}, \textit{volcano}, \textit{goldfish}, \textit{table lamp}, \textit{tram}, \textit{palace}, \textit{tower}, \textit{cup}, \textit{desk}, \textit{chair}, \textit{pot}, \textit{laptop}, \textit{door}, and \textit{car}. For quantitative comparisons, we utilize object classes in CIFAR100~\cite{krizhevsky2009learning}, i.e., 100 classes, as the target objects. The details of the objects are presented in~\cref{tab:supp_obj_stage2}, where the 100 classes are categorized into 20 superclasses for better visualization.

% \newpage
% \clearpage
\subsection{More Qualitative Results}\label{sup:more_qualitative_results}
\noindent\textbf{Stylized generation on various styles.}
We provide additional visualization results in~\cref{fig:sup_figure_elephant_tank,fig:sup_figure_vangogh_and_water}.
Specifically, ~\cref{fig:sup_figure_elephant_tank} show qualitative results on various reference styles. ~\cref{fig:sup_figure_vangogh_and_water} shows qualitative results on various target objects. 
On the one hand, our method can capture fine-grained style details as well as generate high-quality objects, demonstrating the effectiveness of our approach. On the other hand, our method can be utilized to generate various stylized objects, indicating the universality of our proposed method.

\noindent\textbf{Comparison results.}
In addition to~\cref{fig:figure_compare,fig:figure_styletransfer_compare} in the main paper, we provide more comparison results with other methods to further show the superiority of our method in~\cref{fig:sup_figure_compare_a_b,fig:sup_figure_compare_c_d,fig:sup_figure_stytransfer}.
As present in ~\cref{fig:sup_figure_compare_a_b,fig:sup_figure_compare_c_d}, our \textsc{InstaStyle} exhibits better performance in style preservation and content generation. Take~\cref{fig:sup_figure_compare_a_b} (a) as an example, our \textsc{InstaStyle} achieves better image quality than StyleDrop. Although DreamBooth can preserve the style of the reference image, generating target objects is challenging for DreamBooth. Custom Diffusion and Textual Inversion can generate objects in high fidelity. However, the generated images are in the style of a superclass style as the reference image, rather than the fine-grained style of the reference image. \cref{fig:sup_figure_stytransfer} presents comparison results with style transfer methods. Despite the inherent challenge of utilizing text as content input in contrast to style transfer methods that use the content image as input, our \textsc{InstaStyle} demonstrates comparable performance in content fidelity. Additionally, our approach outperforms style transfer methods in preserving the style details in the reference image.

% fig:sup_figure_compare_a,fig:sup_figure_compare_c,
\noindent\textbf{Combination of multiple styles.}
In addition to~\cref{fig:figure_style_combine,fig:figure_style_combine_scale} in the main paper, we provide additional style combination visualization results in~\cref{fig:sup_figure_stysty_2,fig:sup_figure_stysty_2_2}.
We set the noise mix ratio $\alpha$ and the prompt mix ratio $\beta$ to be equal, which are 0, 0.1, 0.3, 0.5, 0.7, 0.9, and 1 from left to right.
As shown in~\cref{fig:sup_figure_stysty_2}, our approach supports adjusting the degree of two styles during combination and can generate various target objects,  demonstrating the flexibility and universality of our approach.
To better illustrate the creative ability of our method, we present the style combination results of a fixed style with different other styles in~\cref{fig:sup_figure_stysty_2_2}.
As shown in~\cref{fig:sup_figure_stysty_2_2}, when the mix ratio is small, \eg, 0.1, the generation results among different style combinations look similar. This can be attributed to the information from the second style is too little, which is not enough to affect the style of the generated image. When the mix ratio is medium, \eg, 0.5, the results present better style combination effects and there is a significant difference among different style combinations. For example, the results in Fig.~\ref{fig:sup_figure_stysty_2_2} (a) contain green color and textures from the second style. The results in Fig.~\ref{fig:sup_figure_stysty_2_2} (b) have some white space just as the reference watercolor dog image. The results in Fig.~\ref{fig:sup_figure_stysty_2_2} (c) have some characteristics of watercolor painting which is similar to the reference watercolor mountain. When the mix ratio is large, 
the resulting images tend towards the second style.

\subsection{More Analysis}
\label{sup:analysis}

\begin{figure}
\begin{center}
\vspace{-6mm}
\includegraphics[width=\linewidth]{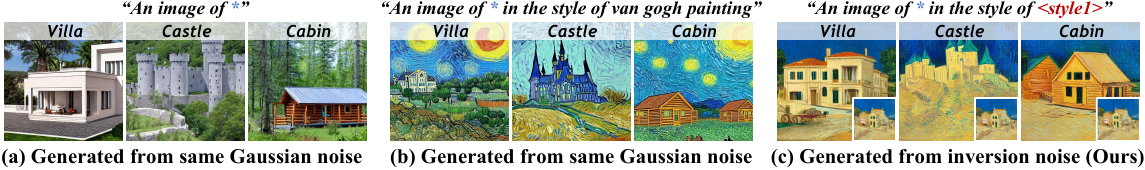}
\end{center}
\vspace{-6mm}
% \vspace{-7mm}
\caption{Visualization of different noise initialization strategy.}\label{fig:initial_noise}
\vspace{-6mm}
% \vspace{-8mm}
\end{figure}

\noindent\textbf{Initial noise.} As our motivation stems from the novel observation that the inversion noise retains style information, we also analyze the effect of the random initialized noise. \cref{fig:initial_noise} (a) and (b) show images generated from the same random Gaussian noise. It can be seen that the output images have some similarities when generating objects with similar characteristics, \eg, different kinds of houses. Mao \etal~\cite{mao2023guided} also observes that images generated from the same Gaussian noise are similar in position and visual appearance, which is consistent with our observations.
\cref{fig:initial_noise} (b) shows that style descriptions can provide some style information, resulting in images in a consistent style. In contrast, as shown in \cref{fig:initial_noise} (c), our method can generate stylized images loyal to the style of the reference image with the help of the inversion noise.

\begin{figure}
\begin{center}
% \vspace{-3.5mm}
\vspace{-6mm}
\includegraphics[width=0.7\linewidth]{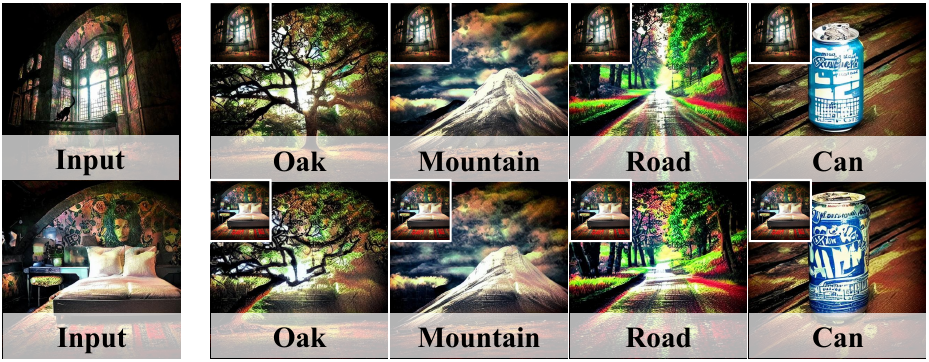}
\end{center}
% \vspace{-7mm}
\vspace{-6mm}
\caption{Results generated from images with different content.} 
\label{fig:init_noise}
% \vspace{-5mm}
\vspace{-6mm}
\end{figure}

In Fig.~\ref{fig:init_noise}, we present generated results from inverted 
noise of images with the same style and different content, revealing that the leaked signal consistently contains the style throughout inverted images.

\begin{figure}
\begin{center}
\vspace{-4mm}
\includegraphics[width=0.6\linewidth]{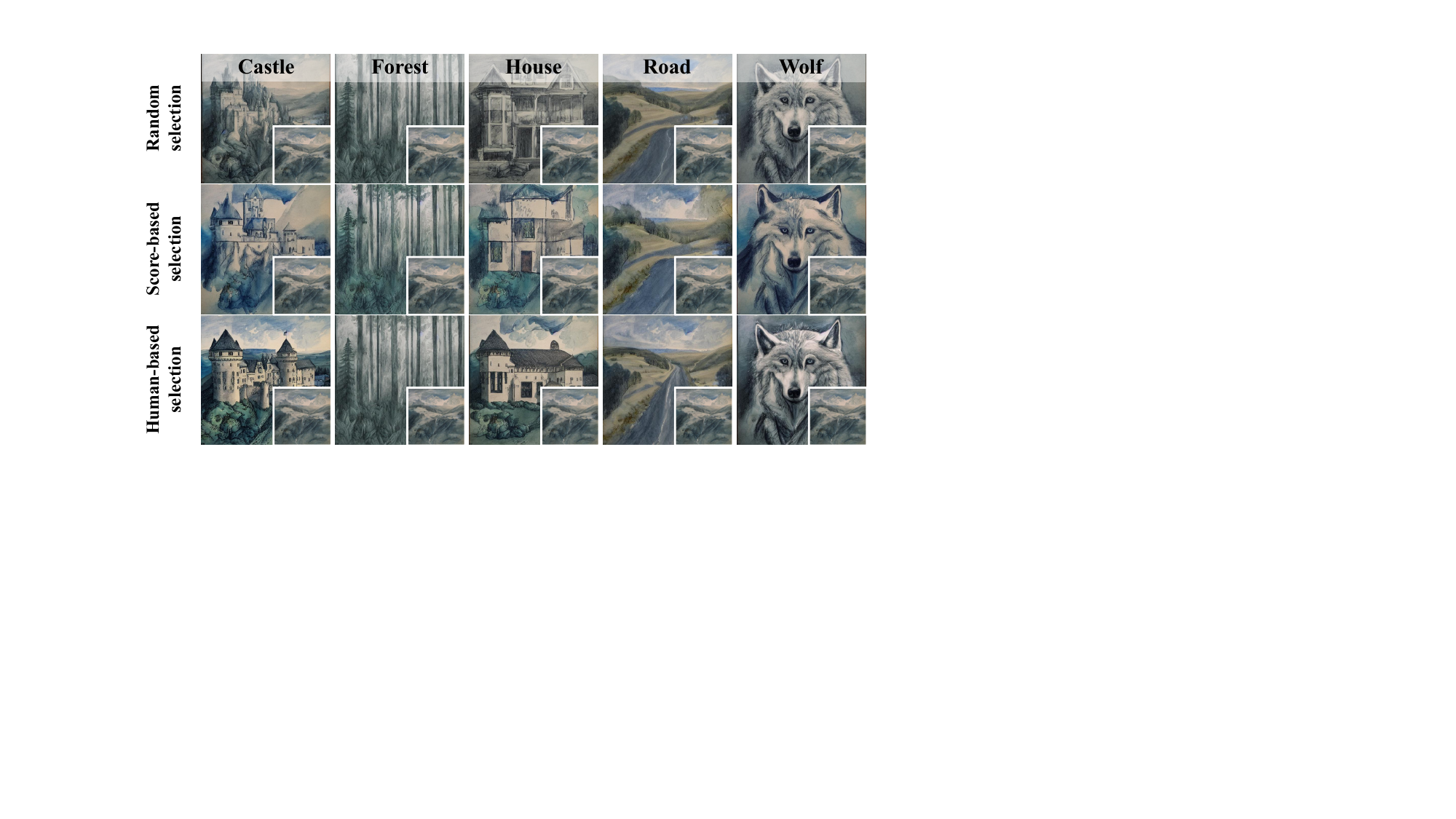}
\end{center}
\vspace{-4mm}
\caption{Qualitative results of selection approaches.}\label{fig:selection}
\vspace{-4mm}
\end{figure}

\noindent\textbf{Selection approaches.}
We experiment with various strategies for selecting high-quality images:
(1) Random selection: we choose several images randomly; (2) Score-based selection: as the style consistency score and content fidelity score are different in scale, we convert them into style rank ($Rank_s$) and content rank ($Rank_c$) by sorting the generated images based on style consistency score and content fidelity score, respectively. Then, the overall rank of an image is calculated as $Rank=max(Rank_s, Rank_c)$. 
(3) Human selection: we manually select images that distinctly embody the reference style and the target object. 
As shown in~\cref{tab:selection},
the score-based selection achieves better results than random selection by considering the content score and style score simultaneously. Our human selection shows the best performance as the selected images are more consistent with human preference.
In order to more intuitively show the comparison between the various methods, we also provide qualitative comparison results in Fig.~\ref{fig:selection}.
The results show that the score-based selection ensures adequate fidelity, making it an effective alternative strategy.

\begin{table}
\vspace{-4mm}
    \centering
    \caption{ Quantitative comparison of different selection approaches.
    }
    \label{tab:selection}
    % \vspace{-4mm}
    \resizebox{0.6\linewidth}{!}{%
    \setlength{\tabcolsep}{2mm}{
    \begin{tabular}{lcc}
    \toprule
    \textbf{Method} &Style ($\uparrow$) &Content ($\uparrow$)\\

    \midrule
    \textsc{InstaStyle}+Random &0.618 &0.285\\
    \textsc{InstaStyle}+Score &0.627 &0.289\\
    \textbf{\textsc{InstaStyle}+Human} &\textbf{0.655} &\textbf{0.294}\\

    \bottomrule
    \end{tabular}
    }}
    \vspace{-4mm}
    % \vspace{-2em}
\end{table}

\begin{figure}
\begin{center}
\includegraphics[width=\linewidth]{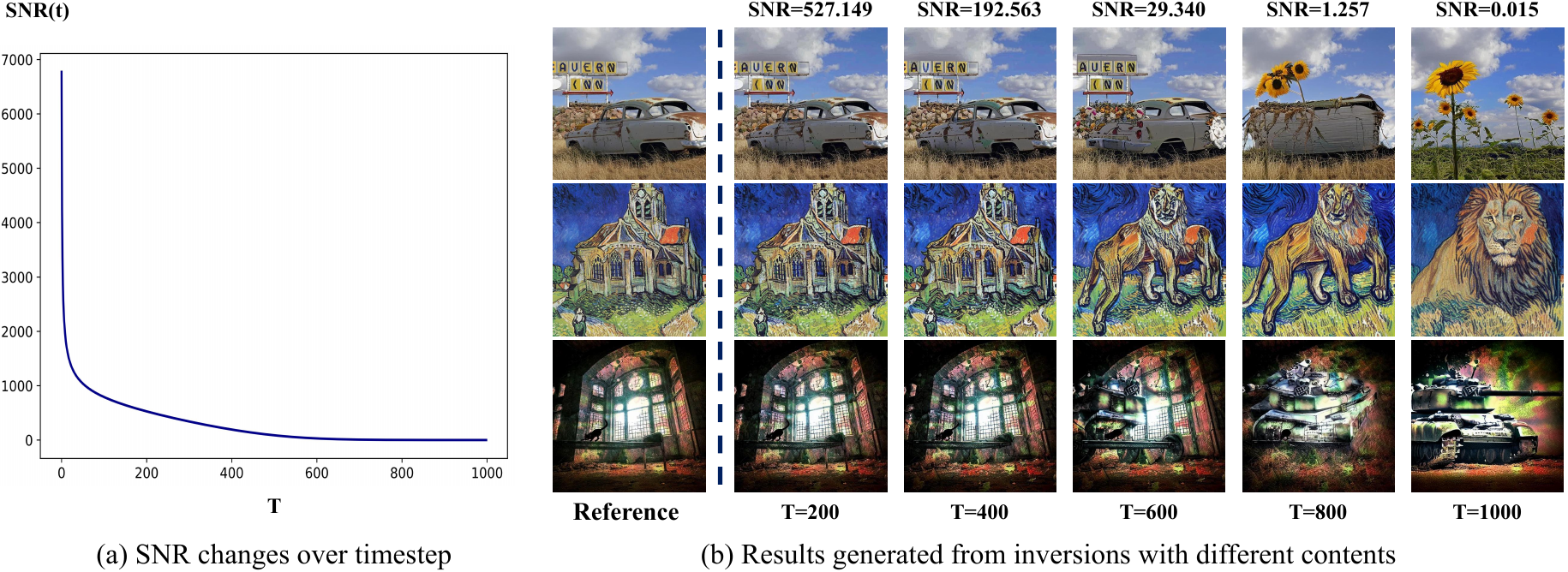}
\end{center}
\vspace{-4mm}
\caption{\textbf{Analysis of the timestep T.} (a) With the increase of T, the SNR decreases. (b) With the increase of T, the style information of the picture can always be preserved, while the redundant structural information, \eg, the object in the reference image, is gradually eliminated. Objects for synthesis are \textit{Sunflowers} and \textit{Lion}.} \label{fig:timestep}
\vspace{-2mm}
\end{figure}

\noindent\textbf{Analysis of the timestep of DDIM inversion.} \label{sup:analysis_DDIM}
We analyze the effect of the terminal timestep of DDIM inversion in inference time. 
As shown in Fig.~\ref{fig:timestep} (a), SNR decreases as T increases.
As shown in Fig.~\ref{fig:timestep} (b), with the increase of timestep, the style information of the picture can always be preserved, while the redundant structural information, \eg, the object in the reference image, is gradually eliminated. On the one hand, this further demonstrates our finding that the inversion noise from a stylized reference image inherently carries the style signal. On the other hand, the avoiding of redundant structural information makes our approach flexible to generate new content.
% although a smaller terminal timestep can result in an inversion image with more information of the reference image, the generation ability is decreased due to the redundant structural information in noise.
In practice, we set the timestep as 1000 for a trade-off between preserving the style information and avoiding the negative effects of redundant information.

\begin{figure}[h]
\begin{center}
\includegraphics[width=\linewidth]{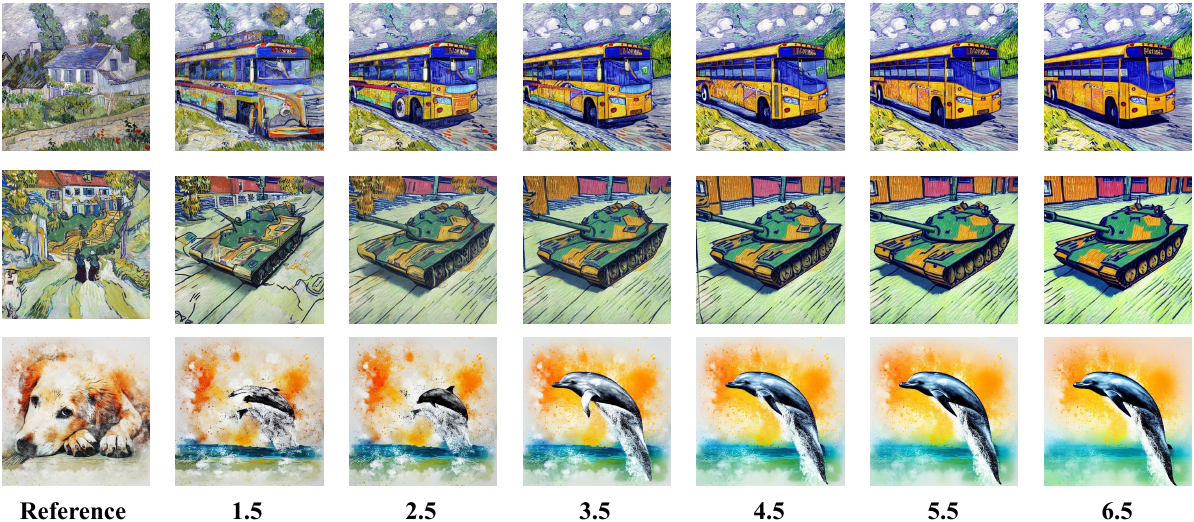}
\end{center}
% \vspace{-4mm}
\caption{\textbf{Visualization of various guidance scales.} A medium guidance scale can make a trade-off between the style and content. The guidance scale in inference is set to 1.5, 2.5, 3.5, 4.5, 5.5, and 6.5, respectively. Objects for synthesis are \textit{Bus} and \textit{Tank}.}\label{fig:scale}
% \vspace{-2mm}
\end{figure}

\noindent\textbf{Analysis of guidance scale.} 
In Fig.~\ref{fig:scale}, we provide visualization results of \textsc{InstaStyle} with varying levels of guidance scales.
When the guidance scale is small, the generated image can preserve the style information in the reference image but may have difficulty in generating the target object. As the guidance scale increases, the model synthesizes more precise and refined objects at the price of losing the style information. A medium guidance scale can make a trade-off between the style and content and we set the guidance scale to 2.5.

\section{Limitations}
\begin{figure}
\begin{center}
% \vspace{-3.5mm}
\includegraphics[width=\linewidth]{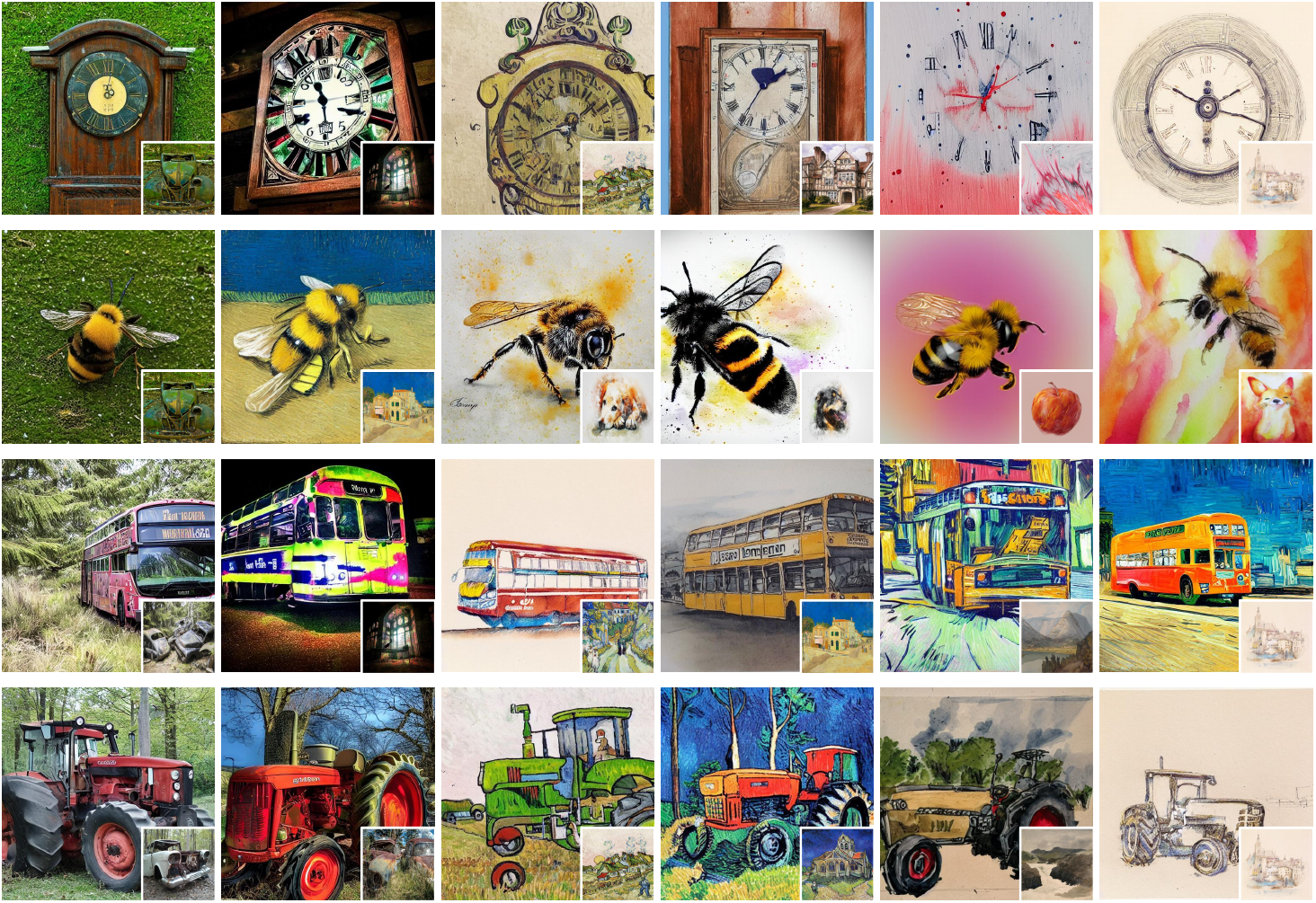}
\end{center}
% \vspace{-7mm}
\caption{\textbf{Limitations of our method.} Our limitation lies in the generation of fine details within target objects, such as the digits on a clock face, the wings of a bee, the wheels of a bus, and some complex mechanical structures on a tractor. }\label{fig:bad_case}
% \vspace{-5mm}
\end{figure}

In \cref{fig:bad_case}, we present instances where our approach encounters challenges. A notable limitation lies in the intricate generation of fine details within target objects, such as the digits on a clock face, the wings of a bee, the wheels of a bus, and some complex mechanical structures on a tractor. 
It is a common challenge in image generation~\cite{zhou2023enhancing}. This limitation might be solved by using a more powerful diffusion model, or by optimizing the inversion noise to better exploit the style information. We will explore it in the future.

